\documentclass[10pt,twocolumn,letterpaper]{article}

\usepackage{cvpr}              % To produce the CAMERA-READY version
\usepackage[dvipsnames]{xcolor}

\definecolor{cvprblue}{rgb}{0.21,0.49,0.74}
\usepackage[pagebackref,breaklinks,colorlinks,citecolor=cvprblue]{hyperref}
\usepackage[accsupp]{axessibility}
\usepackage{amsthm}
\usepackage{bbm}
\usepackage{booktabs}
\theoremstyle{definition}
\usepackage{multirow}
\newtheorem{definition}{Definition}
\def\paperID{33} % *** Enter the Paper ID here
\def\confName{CVPR}
\def\confYear{2024}

\title{Calibration of Continual Learning Models}

\author{
    Lanpei Li\textsuperscript{$\ast$}\\
    University of Pisa and ISTI-CNR\\
    \href{mailto:lanpei.li@isti.cnr.it}{\tt\small lanpei.li@isti.cnr.it}
\and
    Elia Piccoli\textsuperscript{$\ast$}\\
    University of Pisa\\
    \href{mailto:elia.piccoli@phd.unipi.it}{\tt\small elia.piccoli@phd.unipi.it}
\and
    Andrea Cossu\\
    University of Pisa\\
    \href{mailto:andrea.cossu@di.unipi.it}{\tt\small andrea.cossu@di.unipi.it}
\and
    Davide Bacciu\\
    University of Pisa\\
    \href{mailto:davide.bacciu@unipi.it}{\tt\small davide.bacciu@unipi.it}
\and
    Vincenzo Lomonaco\\
    University of Pisa\\
    \href{mailto:vincenzo.lomonaco@unipi.it}{\tt\small vincenzo.lomonaco@unipi.it}
}

\newcommand{\customfootnotetext}[2]{{% Group to localize change to footnote
  \renewcommand{\thefootnote}{#1}% Update footnote counter representation
  \footnotetext[0]{#2}}}% Print footnote text

\begin{document}
\maketitle

\customfootnotetext{$\ast$}{Corresponding authors.}

\begin{abstract}
    Continual Learning (CL) focuses on maximizing the predictive performance of a model across a non-stationary stream of data. Unfortunately, CL models tend to forget previous knowledge, thus often underperforming when compared with an offline model trained jointly on the entire data stream. Given that any CL model will eventually make mistakes, it is of crucial importance to build \emph{calibrated} CL models: models that can reliably tell their confidence when making a prediction. Model calibration is an active research topic in machine learning, yet to be properly investigated in CL. We provide the first empirical study of the behavior of calibration approaches in CL, showing that CL strategies do not inherently learn calibrated models. To mitigate this issue, we design a continual calibration approach that improves the performance of post-processing calibration methods over a wide range of different benchmarks and CL strategies. CL does not necessarily need perfect predictive models, but rather it can benefit from reliable predictive models. We believe our study on continual calibration represents a first step towards this direction. 
\end{abstract}

\section{Introduction}
\begin{figure*}[t]
\centering
\includegraphics[width=0.9\linewidth]{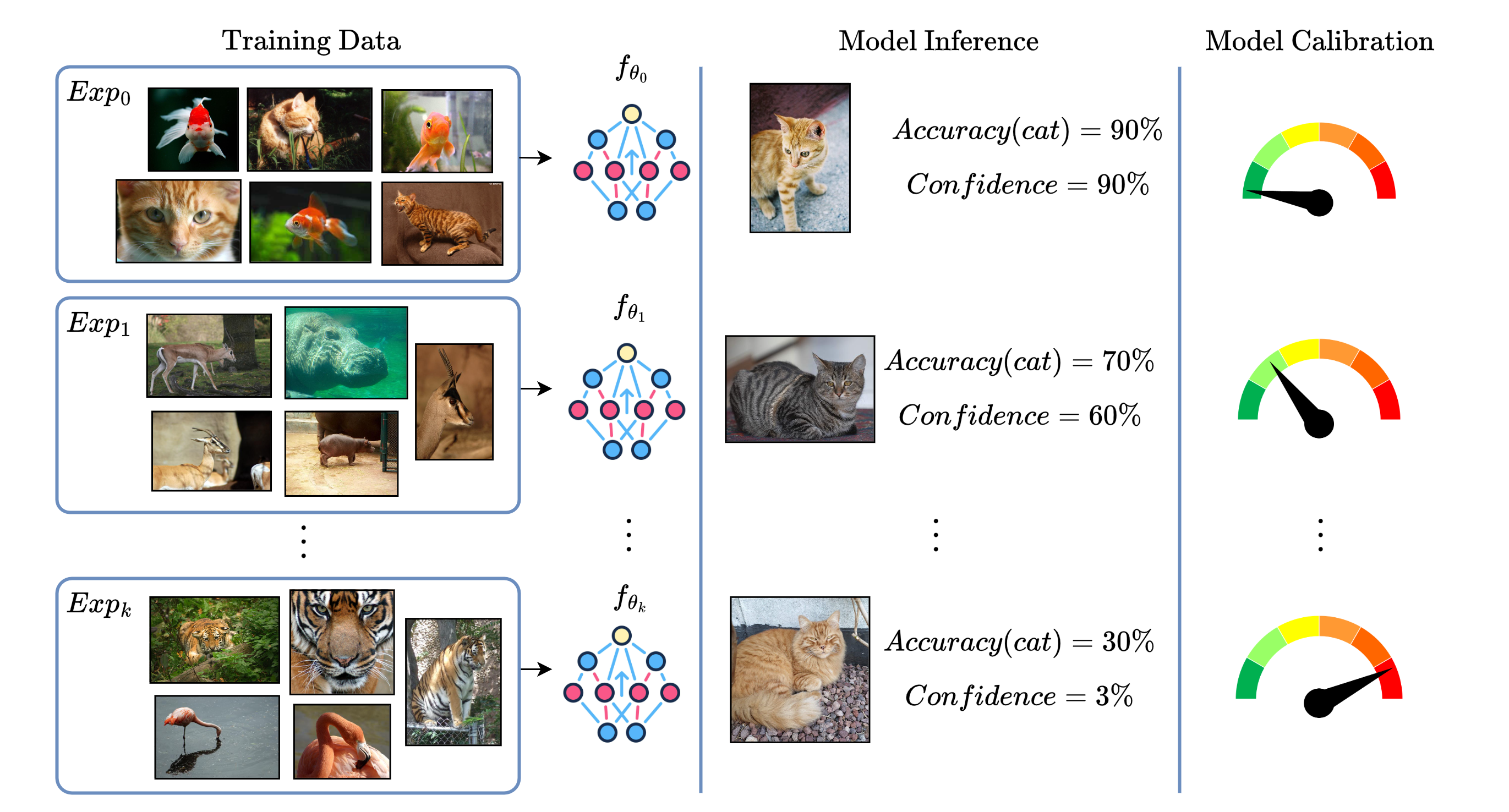}
    \caption{A CL model $f_{\theta}$ is trained on a sequences of $k$ experiences (or tasks). The model accuracy on the class ``cat'' decreases over time. Its confidence decreases much faster. Therefore, the model becomes less calibrated over the course of its learning phase. A calibrated CL model, which is the objective of this paper, should output a confidence which is equal to the average accuracy. A calibrated model knows what to expect, on average, as a result of its predictions.}
    \label{fig:why_cc}
\end{figure*}

In offline machine learning, models learn from a fixed data distribution and they are tested on new examples from the same distribution (the \emph{iid} assumption). Unfortunately, machine learning models never achieve perfect predictive accuracy unless the task is very simple or created ad-hoc. \\
If a perfect predictive model is unrealistic for offline machine learning, it is even more unlikely in Continual Learning (CL) \cite{lesort2020}, where the model learns over time from a sequence of non-stationary data distributions. Due to the forgetting phenomenon \cite{french1999}, the predictive performance of a CL model can degrade as the model is trained on new distributions.\\
So far, most of the efforts in CL have been dedicated to designing approaches that mitigate forgetting and increase predictive performance \cite{delange2021}. While these remain fundamental challenges for the advancement of CL research, they also mainly aim at reducing the gap with respect to a perfect predictive model. We do not expect this gap to be ever fully closed, hence \emph{we need to learn how to deal with imperfect models that make mistakes}.\\

The objective of this paper is to understand how to build CL systems that can be trusted. For example, being able to tell \emph{in advance} when a model might be wrong can make applications more robust and reliable: a user could discard predictions that do not match a predefined level of trust and only accept those that are marked as safe by the model itself (as in the learning to reject paradigm \cite{cortes2016}). To this extent, we leverage the \emph{calibration} paradigm, a well-known research topic in machine learning that aims at learning a proper \emph{confidence} measure related to the predictions of a model \cite{zhang2023, guo2017, pereyra2017a}. \\
Intuitively, the confidence tells how likely the model is, on average, to provide a correct answer on a given example. For this reason, calibrated models are extremely useful in many practical scenarios, from finance and healthcare to computer vision and robotics. The more autonomous the application, the more risky it is to rely on the predictions of uncalibrated models. Although it could be of extreme use for practical purposes, calibration is currently disregarded in CL (with the notable exception of a brief mention in \cite{buzzega2020, boschini2022a}, that we also consider in our work ).\\
% {\color{red} and some insights concerning distribution shifts between training and test data in \cite{minderer21revisiting}}
We believe calibration to be a fundamental challenge for CL as it is unlikely to achieve reliable CL models for real-world applications without a strong notion about their robustness (Figure \ref{fig:why_cc}). Here, we provide our contribution towards calibrated CL models:
\begin{enumerate}
    \item We ran extensive experiments across 4 CL benchmarks, 3 CL strategies, and 5 calibration methods. We include both popular CL benchmarks as well as others datasets aimed at testing calibration in scenarios beyond computer vision (supervised action prediction from Atari) and real-world scenarios (land use detection from satellite images). To the best of our knowledge, this is the first comprehensive study on continual calibration. 
    \item We discovered that calibration methods only partially work when applied to non-stationary data streams. Even when equipped with CL strategies, the resulting models are not necessarily well calibrated, especially when compared with the same model trained offline on the entire data stream. 
    \item We design Replayed Calibration, a continual calibration method that is compatible with a large family of calibration approaches (the post-processing calibration approaches introduced in Section \ref{sec:methods}). Our approach improves the performance of calibration methods by large margins.
\end{enumerate}

\section{Calibration background} \label{sec:background}
Calibration has been studied for the offline machine learning setup \cite{zhang2023, vaicenavicius2019}. We provide a brief overview of calibration mainly intended for continual learning researchers who are interested in applying or studying calibration. \\
We focus on the calibration of neural network models trained on supervised classification tasks \cite{guo2017}. Most of this discussion generalizes to other types of models as well. \\

A model $f_\theta$ parameterized by $\theta \in \mathbb{R}^d$ is trained on a dataset $\mathcal{D} = \{(x_j, y_j)\}_{j=1,\ldots,M}$, where each example is composed by an input-target pair $(x_j, y_j)$ and the target represents the class associated with the input. For each input example $x_j$ the model returns a probability vector $\hat{y}_j$, containing one probability per class, \emph{and} a confidence value $\hat{c}_j \in \mathbb{R}$. Formally, $\hat{y}_j, \hat{c}_j = f_\theta(x_j)$. The probability vector is obtained by passing the logits $z_j$ through a softmax function: $\hat{y}_j = \text{softmax}(z_j)$. The logits are computed by the last layer of the model $z_j = h_\theta(x_j)$, where $h_\theta$ computes the pre-softmax output. The model is trained by minimizing a loss function $\mathcal{L}(\hat{y}, y)$ (e.g., cross-entropy).
% Equipped with this notation, we can now formally define calibration.
\begin{definition} \label{def:calibration}
A model $f_\theta$ is \emph{calibrated} when $P(\hat{y} = y | \hat{c} = c) = c,\ \forall c \in [0, 1]$.
\end{definition}
Definition \ref{def:calibration} states that a model is calibrated when the probability of predicting the correct class is equal to the confidence, for any given value of the confidence. The calibration objective cannot be computed exactly since the joint distribution $P(\hat{Y}, \hat{C})$ is taken over the predictions and confidence random variables, respectively, which are continuous variables. Given a dataset $\mathcal{D}$ with $M$ examples, we use the Expected Calibration Error (ECE) and the reliability diagrams as approximations of the calibration objective of Definition \ref{def:calibration}. The reliability diagram reports the histogram of accuracy against confidence. The histogram collects all $M$ model predictions and confidence values. Then, it partitions the predictions in $K$ equally-spaced bins based on the confidence value. It finally computes the average accuracy of the predictions within each bin.
\begin{definition} \label{def:reliability}
    A reliability diagram with $K$ equally-spaced confidence bins reports the average accuracy over a generic bin $I_b$ as $\overline{a}_b = \frac{1}{|I_b|} \sum_{c_j \in I_b} \mathbbm{1}(\hat{y}_j = y_j)$. Correspondingly, the average confidence within a bin $I_b$ by $\overline{c}_b = \frac{1}{|I_b|} \sum_{\hat{c}_j \in I_b} \hat{c}_j$.
\end{definition}
A perfectly calibrated model would return a reliability diagram equal to the identity function: $\overline{a}_b = \overline{c}_b,\ \forall I_b$. We use reliability diagrams in our experiments (see Figure \ref{fig:splitcifar100-der} for an example). The distance from a perfectly calibrated model is computed by the Expected Calibration Error, which summarizes the information contained within a reliability diagram in a single value. 
\begin{definition} \label{def:ece}
    The Expected Calibration Error (ECE) for a given model on a dataset $\mathcal{D}$ is computed as $\text{ECE} = \sum_{b=1}^K \frac{|I_b|}{M} | \overline{a}_b - \overline{c}_b|$, where $M$ is the total number of examples in $\mathcal{D}$.
\end{definition}
Notice how ECE is a scalar metric between 0 and 1, hence it can be reported as a percentage value, with $0\%$ representing a perfectly calibrated model and $100\%$ its opposite, not calibrated counterpart.

\subsection{Calibration of neural networks} \label{sec:methods}
Although calibration has been studied for years in machine learning \cite{zhang2023}, there are only a few techniques available that are compatible with neural networks \emph{and} multi-class classification tasks (with more than 2 classes) \cite{guo2017, pereyra2017a}. \\
For this paper, we follow \cite{zhang2023} and divide calibration methods into two main families: \emph{post-processing calibration methods} and \emph{self-calibration methods}. \\
Post-processing calibration methods are applied after the model training phase and they rely on a held-out validation set to tune or learn some calibration (hyper)parameters. The same validation set can also be used for model selection. Many post-processing calibration methods are available for binary classification tasks. Since in a CL environment, new classes often appear over time, it is unrealistic to consider binary classification tasks. Therefore, we focus on existing extensions to the multi-class case. \\
Self-calibration methods operate directly during model training, without requiring a separate calibration phase.

\paragraph{Temperature scaling (TS).} TS \cite{guo2017} is a post-processing calibration method that adapts the softmax temperature applied after the output layer to compute ``softer'' probability distributions. Peaked distributions are often associated with over-confidence in the prediction. TS computes the confidence on an example $x_j$ as $\hat{c}_j = \max \text{softmax}(\frac{z_j}{T})$, where the logits are divided by the scalar temperature $T$ and the maximum is computed across the resulting probability vector after the softmax. The temperature is learned by minimizing the Negative Log Likelihood on the validation set, which is associated with the entropy and therefore measures how peaked the distribution is. Since TS only changes the temperature $T$, the output classes predicted by the model remain the same before and after the calibration phase.

\paragraph{Matrix/Vector scaling.} Matrix scaling (MS) and Vector scaling (VS) \cite{guo2017} are post-processing methods that learn an additional linear projection parameterized by $W, b$ during the calibration phase. The model predictions on a generic example $x_j$ are updated as $\hat{y}_j = \text{softmax}(W z_j + b)$ and the confidence is obtained by $\hat{c}_j = \max \hat{y}_j$ (like TS, the maximum is computed across the probability vector returned by the softmax). The parameters $W, b$ are optimized with respect to the Negative Log Likelihood on the validation set. In MS, $W$ is any matrix, while in VS $W$ is a diagonal matrix (for efficiency purposes).

\paragraph{Entropy regularization (HR).} Instead of promoting high-entropy distributions via post-processing methods like TS and MS/VS, HR \cite{pereyra2017a} operates directly during model training. The loss used at training time is augmented with a regularization term of the form $- \lambda H(\hat{y}_j)$, where $H$ is the entropy of the probability distribution computed by the model on $x_j$. The optimization process strives to minimize the loss, hence to maximize the entropy (and prevent peaked distributions). The regularization is controlled by the hyper-parameter $\lambda$.

\section{Continual Calibration}
\begin{figure}[t]
\centering
\includegraphics[width=\linewidth]{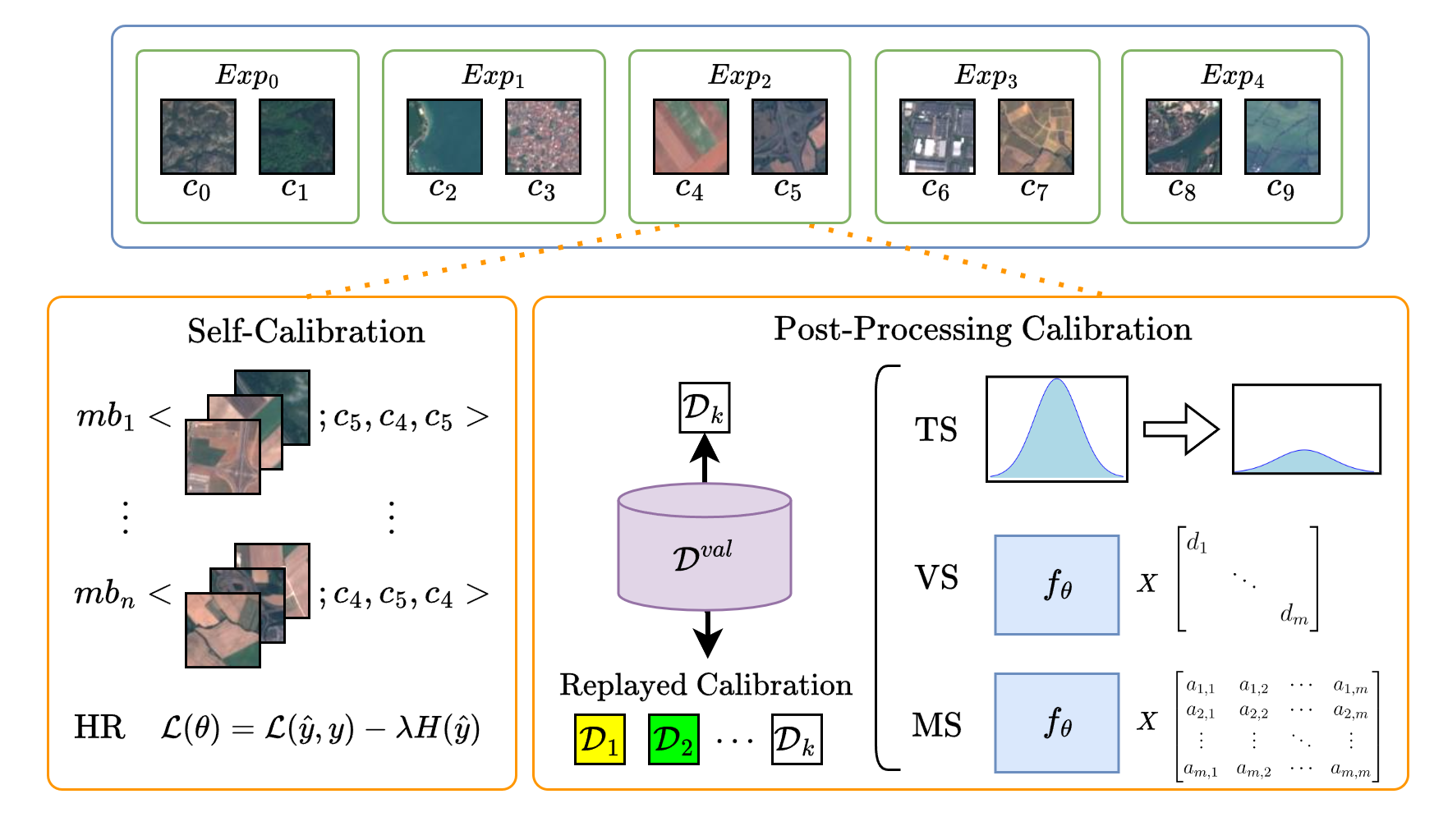}
    \caption{Continual calibration is performed on a stream of experiences (top) by applying either self-calibration (bottom left) or post-processing calibration (bottom right). Self-calibration approaches like Entropy Regularization (HR) regularize the training loss at each minibatch. Post-processing calibration like Temperature Scaling (TS) and Matrix/Vector scaling (MS/VS) are applied only at the end of each experience. Our Replayed Calibration approach is applicable alongside any post-processing methods.}
    \label{fig:cc}
\end{figure}

Our objective is to i) understand how to apply calibration methods in a CL setup, ii) assess the behavior of calibration approaches on non-stationary data streams and iii) extend existing approaches backed by intuitions from CL strategies. Figure \ref{fig:cc} provides a compact representation of all three points. \\

Calibration of CL models is especially challenging, since the data distribution faced during training changes over time. All the calibration methods we presented in Section \ref{sec:methods} are designed for a data distribution that does not change between the training and the calibration phase. In CL, we have a stream of experiences (or tasks) $\mathcal{S} = (e_1, e_2, \ldots)$ \cite{lomonaco2021}. Each experience $e_i$ contains a dataset $\mathcal{D}_i$ with $M_i$ examples: $\mathcal{D}_i = \{(x_j, y_j)\}_{j=1,\ldots,M_i}$. We are still considering the supervised classification setup for CL. The stream $\mathcal{S}$ becomes available over time and the model is continuously trained on each experience sequentially. Importantly, the data distribution changes between one experience and the other, making the stream non-stationary \cite{ditzler2015}. Since the content of each experience cannot be entirely stored for later reuse, the model needs to learn new experiences without forgetting previous ones. That is, the predictive performance on previous experiences should not decrease. \\
Self-calibration techniques like HR are already compatible with a CL setup since they do not require a separate calibration phase. Post-processing calibration methods, instead, only operate \emph{at the end} of the training phase. While this makes sense in an offline learning setup, where all data is available at once, post-processing calibration methods are not directly applicable in CL, where the model could be potentially trained on an infinite sequence of experiences.
\paragraph{Post-processing continual calibration.}
When considering finite data streams, one possibility would be to apply the post-processing calibration method only once at the end of training on all experiences. Unfortunately, since in CL we cannot store the entire content of previous experiences, the post-processing calibration would be applied only to the validation set associated with the last experience. Therefore, the model would not be calibrated on any examples coming from previous experiences. \\
Instead, we add a calibration phase at the end of training on each experience. Calibration is performed on the validation set $\mathcal{D}^{\text{val}}$ associated with the current experience, where $\mathcal{D}_i = \mathcal{D}^{\text{train}}_i \cup \mathcal{D}^{\text{val}}_i$. The test set $\mathcal{D}^{\text{test}}_i$ associated with each experience is assumed to be always available. The test sets are never used neither to train the model nor to calibrate it, but only for evaluation purposes.\\
The post-processing calibration methods we considered either add a new layer after the original classifier (VS, MS) or they change the default softmax temperature from 1 to a learned value (TS). In our CL setup, these changes are first introduced after training on the first experiences. To comply with the CL setup, in the following experiences we did not revert the changes made by post-processing calibration and we train continuously the resulting CL model (either with an extra output layer or with a learned temperature).
\begin{table*}[htbp]
  \centering
  \caption{Average accuracy and standard deviation on the test set of all experiences computed at the end of training. Bold highlights the best result for each CL strategy and Joint Training. Bold and underline highlights overall best across CL strategies.}
  \label{tab:acc}
  \begin{tabular}{c|cccc}
    \toprule
    \textbf{Accuracy (\%)} & \textit{\textbf{Split MNIST}} & \textit{\textbf{Split CIFAR100}} & \textit{\textbf{EuroSAT}} & \textit{\textbf{Atari}}\\
    \midrule
    \textbf{Joint} & $93.00 \pm 1.08$& $64.37 \pm 7.22$ & $91.48 \pm 4.05$& $55.10 \pm 0.61$ \\
    HR & $\mathbf{94.91 \pm 1.10}$ & $62.20 \pm 4.76$ & $94.51 \pm 4.26$& $54.82 \pm 0.71$ \\
    TS & $93.92 \pm 0.71$ & $64.67 \pm 3.18$ & $95.12 \pm 3.75$& $\mathbf{55.38 \pm 0.42}$ \\
    VS & $94.21 \pm 1.76$ & $\mathbf{68.30 \pm 5.92}$ & $93.74 \pm 3.34$& $55.32 \pm 0.30$ \\
    MS & $94.24 \pm 3.32$ & $62.90 \pm 2.90$ & $\mathbf{95.31 \pm 4.69}$& $39.27 \pm 2.04$ \\
    \hline
    \hline
    \textbf{DER++} & $92.74 \pm 0.38$ & $35.18 \pm 2.86$ & $77.39 \pm 8.44$& $32.35 \pm 0.17$ \\
    HR & $92.56 \pm 0.39$ & $37.41 \pm 2.70$ & $79.86 \pm 2.76$& $32.09 \pm 0.34$ \\
    TS & $\underline{\mathbf{94.79 \pm 0.21}}$ & $32.75 \pm 10.43$ & $77.11 \pm 1.21$& $\underline{\mathbf{33.14 \pm 0.67}}$ \\
    VS & $91.92 \pm 0.34$ & $23.30 \pm 2.56$& $70.60 \pm 2.65$& $25.19 \pm 2.75$ \\
    MS & $91.95 \pm 0.19$ & $19.01 \pm 10.98$& $53.20 \pm 18.72$& $25.03 \pm 3.74$ \\
    TS + RC & $94.74 \pm 0.20$ & $\mathbf{41.79 \pm 0.84}$ & $\mathbf{80.81 \pm 4.34}$& $33.03 \pm 0.87$ \\
    VS + RC & $92.24 \pm 0.27$ & $34.26 \pm 6.08$& $57.75 \pm 21.32$& $26.21 \pm 1.42$ \\
    MS + RC & $92.23 \pm 0.02$ & $37.30 \pm 5.82$& $75.07 \pm 3.38$& $25.56 \pm 2.16$ \\
    \hline
    \textbf{Replay} & $90.97 \pm 0.66$ & $40.39 \pm 4.00$ & $80.57 \pm 1.06$& $29.30 \pm 0.66$ \\
    HR & $90.71 \pm 0.63$ & $40.36 \pm 1.38$ & $81.03 \pm 0.74$& $\mathbf{29.87 \pm 0.50}$ \\
    TS & $\mathbf{94.20 \pm 0.55}$ & $23.22 \pm 7.56$ & $78.02 \pm 0.67$& $28.29 \pm 0.51$ \\
    VS & $75.31 \pm 2.31$& $35.00 \pm 2.90$ & $73.08 \pm 2.39$& $28.45 \pm 0.19$ \\
    MS & $74.18 \pm 5.54$& $34.73 \pm 3.33$ & $79.37 \pm 4.43$& $28.98 \pm 0.38$ \\
    TS + RC & $93.87 \pm 0.81$ & $15.26 \pm 8.59$ & $81.73 \pm 0.76$& $29.03 \pm 0.42$ \\
    VS + RC & $90.19 \pm 0.85$& $\underline{\mathbf{46.82 \pm 0.87}}$ & $\underline{\mathbf{84.95 \pm 2.01}}$& $27.83 \pm 0.93$ \\
    MS + RC & $90.77 \pm 0.32$& $46.42 \pm 0.93$ & $83.72 \pm 0.91$& $28.34 \pm 0.76$ \\
    \hline
    \textbf{Naive} & $19.86 \pm 0.07$& $7.90 \pm 0.65$ & $19.84 \pm 0.22$& $20.57 \pm 1.76$ \\
    HR & $20.46 \pm 0.45$& $8.29 \pm 0.70$ & $19.36 \pm 0.35$& $19.95 \pm 0.49$ \\
    TS & $21.40 \pm 0.35$& $7.97 \pm 0.58$ & $19.58 \pm 0.41$& $20.96 \pm 0.94$ \\
    VS & $34.85 \pm 0.05$& $7.98 \pm 0.24$ & $19.90 \pm 0.17$& $20.66 \pm 0.62$ \\
    MS & $19.58 \pm 0.31$& $8.10 \pm 0.44$ & $19.41 \pm 0.16$& $19.49 \pm 0.25$ \\
    TS + RC & $19.72 \pm 0.16$& $8.21 \pm 0.26$ & $19.71 \pm 0.40$& $21.40 \pm 0.02$ \\
    VS + RC & $34.85 \pm 11.13$& $10.57 \pm 0.38$ & $15.32 \pm 5.15$& $20.87 \pm 1.01$ \\
    MS + RC & $\mathbf{37.80 \pm 5.26}$ & $\mathbf{11.96 \pm 1.03}$ & $\mathbf{19.93 \pm 1.87}$& $\mathbf{21.96 \pm 1.06}$ \\
    \bottomrule
  \end{tabular}
\end{table*}
\begin{table*}[htbp]
  \centering
  \caption{Average ECE (10 bins) and standard deviation on the test set of all experiences computed at the end of training. Bold highlights the best result for each CL strategy and Joint Training. Bold and underline highlights overall best across CL strategies.}
  \label{tab:ece}
  \begin{tabular}{c|cccc}
    \toprule
    \textbf{ECE (\%)} & \textit{\textbf{Split MNIST}} & \textit{\textbf{Split CIFAR100}} & \textit{\textbf{EuroSAT}} & \textit{\textbf{Atari}}\\
    \midrule
    \textbf{Joint} & $4.53 \pm 0.40$ & $14.60 \pm 6.23$ & $4.45 \pm 2.13$& $2.20 \pm 1.83$\\
    HR & $2.85 \pm 1.04$& $15.91 \pm 3.00$ & $3.12 \pm 3.34$& $1.90 \pm 0.79$\\
    TS & $\mathbf{1.56 \pm 0.32}$& $7.80 \pm 4.12$ & $2.57 \pm 1.78$& $1.52 \pm 0.71$ \\
    VS & $1.70 \pm 0.27$& $\mathbf{5.75 \pm 2.03}$ & $\mathbf{2.46 \pm 1.81}$& $\mathbf{1.38 \pm 0.20}$ \\
    MS & $38.08 \pm 2.05$& $27.96 \pm 4.71$ & $39.67 \pm 2.97$& $23.50 \pm 2.97$ \\
    \hline
    \hline
    \textbf{DER++} & $1.96 \pm 0.42$ & $36.45 \pm 4.09$ & $11.29 \pm 1.67$& $12.51 \pm 0.65$ \\
    HR & $1.96 \pm 0.20$ & $38.70 \pm 1.33$ & $10.69 \pm 0.78$& $11.69 \pm 1.64$ \\
    TS & $\underline{\mathbf{1.25 \pm 0.13}}$ & $28.42 \pm 2.42$ & $13.00 \pm 2.68$& $7.25 \pm 0.29$ \\
    VS & $4.86 \pm 0.35$ & $34.64 \pm 0.59$& $17.93 \pm 4.07$& $7.67 \pm 1.97$ \\
    MS & $4.08 \pm 0.16$ & $37.27 \pm 0.29$& $28.75 \pm 8.81$& $8.98 \pm 4.33$ \\
    TS + RC & $1.43 \pm 0.13$ & $27.52 \pm 1.96$ & $\mathbf{9.00 \pm 5.13}$& $6.13 \pm 0.69$ \\
    VS + RC & $3.63 \pm 0.77$ & $9.59 \pm 1.16$& $15.45 \pm 8.85$& $4.69 \pm 0.49$ \\
    MS + RC & $3.77 \pm 0.26$ & $\underline{\mathbf{8.51 \pm 0.67}}$& $10.86 \pm 2.90$& $\underline{\mathbf{3.61 \pm 1.16}}$ \\
    \hline
    \textbf{Replay} & $3.77 \pm 0.32$& $38.42 \pm 5.95$ & $11.07 \pm 2.09$& $48.94 \pm 0.68$ \\
    HR & $3.85 \pm 0.35$& $39.31 \pm 1.76$ & $9.46 \pm 0.71$& $48.68 \pm 1.24$ \\
    TS & $2.86 \pm 0.69$& $23.72 \pm 4.35$ & $12.96 \pm 2.65$& $49.96 \pm 2.28$ \\
    VS & $8.21 \pm 3.05$& $50.11 \pm 2.84$ & $15.06 \pm 0.22$& $34.78 \pm 5.09$ \\
    MS & $8.99 \pm 4.26$& $49.76 \pm 1.82$ & $12.91 \pm 3.53$& $41.13 \pm 2.62$ \\
    TS + RC & $\mathbf{2.24 \pm 0.25}$& $\mathbf{14.63 \pm 3.35}$ & $8.23 \pm 0.96$& $41.12 \pm 1.87$ \\
    VS + RC & $4.29 \pm 0.62$& $26.04 \pm 1.25$ & $\underline{\mathbf{4.70 \pm 0.64}}$& $13.75 \pm 1.23$ \\
    MS + RC & $3.76 \pm 0.45$& $27.12 \pm 1.61$ & $9.46 \pm 0.62$& $\mathbf{13.46 \pm 3.40}$ \\
    \hline
    \textbf{Naive} & $70.31 \pm 0.79$& $72.44 \pm 3.15$ & $76.57 \pm 1.61$& $34.60 \pm 5.35$ \\
    HR & $67.05 \pm 3.95$& $70.40 \pm 3.53$ & $74.88 \pm 2.25$& $34.51 \pm 4.20$ \\
    TS & $71.94 \pm 2.08$& $67.07 \pm 2.24$ & $76.31 \pm 0.55$& $47.10 \pm 3.33$ \\
    VS & $73.64 \pm 0.84$& $65.89 \pm 1.07$ & $78.17 \pm 0.31$& $26.35 \pm 8.56$ \\
    MS & $72.73 \pm 4.09$& $63.56 \pm 1.81$ & $75.70 \pm 2.64$& $35.77 \pm 5.71$ \\
    TS + RC & $65.57 \pm 2.31$& $61.26 \pm 2.99$ & $74.58 \pm 1.24$& $42.03 \pm 4.61$ \\
    VS + RC & $30.31 \pm 5.75$& $25.73 \pm 1.29$ & $35.69 \pm 4.99$& $29.20 \pm 3.00$ \\
    MS + RC & $\mathbf{28.07 \pm 3.40}$& $\mathbf{21.99 \pm 1.35}$ & $\mathbf{31.27 \pm 2.27}$& $\mathbf{13.87 \pm 5.29}$ \\
    \bottomrule
  \end{tabular}
\end{table*}

\paragraph{Replayed Calibration (RC).}
Our adaptation of post-processing calibration for CL does not completely solve the issue of calibrating on an incomplete portion of the data. During each calibration phase, the model only sees data coming from the current experience. Therefore, when a new experience arrives (a new data distribution), we have no guarantee that the previously calibrated model will remain calibrated on previous distributions. We extend post-processing calibration methods with CL approaches based on replay \cite{hayes2021}. Many CL applications allow to store a (small) subset of previous data. Usually, replay techniques leverage the external buffer at training time by training the model on data coming from the current experience and from the memory buffer, to improve model stability and mitigate forgetting. Inspired by this approach, we do the same during the calibration phase. The external memory buffer contains examples from the validation sets of previous experiences. The model is then calibrated on both the content of the buffer \emph{and} the validation set of the current experience. We call this post-processing calibration approach \emph{Replayed Calibration (RC)}. RC can be combined with any of the existing post-processing calibration methods.

\subsection{Empirical evaluation}
We study calibration of CL models trained with Naive finetuning, Experience Replay \cite{rolnick2019} and Dark Experience Replay (DER) \cite{buzzega2020}, in its DER++ version\footnote{The code to reproduce the experiments is available at \url{https://github.com/lilanpei/Continual-Calibration} and as supplementary material.}. Naive simply trains the model continuously over the data stream, minimizing the classification loss. Experience Replay keeps a fixed-size buffer in which to store examples from previous experiences. We use reservoir sampling to fill the buffer. DER++ is a state-of-the-art CL method that combines replay and distillation. In addition to input-target pairs, the replay memory $M$ of DER++ also stores the logits computed by the model when the example was first added to the memory. The distillation loss is computed on examples sampled from the memory and it reads $\alpha\ \mathbb{E}_{(x, z) \sim M} \| z - h(x) \|^2_2 + \beta\ \mathbb{E}_{(x', y') \sim M} \mathcal{L}(\hat{y}, y')$, where for simplicity $\hat{y}$ in the last term denotes the model prediction computed on $x'$. DER++ uses two hyper-parameters $\alpha$ and $\beta$ to control the contribution of each regularizer. Intuitively, the regularizer controlled by $\alpha$ promotes stability of the output distribution, while the regularizer controlled by $\beta$ prevents a drop in the predictive performance on previous examples (since $\mathcal{L}$ is the classification loss).\\
Interestingly, DER++ is known to result in calibrated models. However, the original paper \cite{buzzega2020, boschini2022a} did not consider any calibration methods. We combined DER++ with $5$ calibration methods, including our RC and we verified whether we can improve DER++ calibration.\\
We compare all the methods with the offline learning model jointly trained on the dataset resulting from the concatenation of all experiences: $\mathcal{D} = \cup_{i=1}^N e_i$, for a stream with $N$ experiences. Ideally, we would like CL models to achieve a similar calibration than the offline learning models. As expected, due to the continuous training and the presence of drifts between experiences the CL models under-perform with respect to the offline models. All CL strategies are coupled with various calibration methods, including self-training HR, three post-processing calibration techniques (TS, VS, and MS), and our Replayed Calibration RC.

\paragraph{Benchmarks.}
We assess the performance of calibration methods on 4 CL benchmarks: Split MNIST \cite{vandeven2018}, Split CIFAR100 \cite{rebuffi2017, lopez-paz2017}, EuroSAT \cite{helber2018introducing,helber2019eurosat} and Atari \cite{bellemare13arcade, machado18arcade}. Split MNIST is obtained by splitting the MNIST dataset into 5 experiences, each of which contains examples from 2 classes. Similarly, Split CIFAR100 is obtained by splitting the CIFAR100 dataset into 10 experiences, each of which contains examples from 10 classes. Both benchmarks are class-incremental benchmarks \cite{rebuffi2017}.\\
% Split MNIST and Split CIFAR100 are very popular benchmarks.
EuroSAT is a publicly available dataset for land use and land cover classification from Sentinel-2 satellite images. We adopted this dataset as it represents an interesting example of a CL application in a resource-constrained environment. The CL agent can operate directly on the satellite in an autonomous way. Therefore, it needs to provide robust, calibrated predictions.
% where access to online resources (e.g., Internet) may be restricted or prevented and where the onboard memory is limited.
We created a class-incremental benchmark by splitting the dataset into 5 experiences, each of which contains JPEG-encoded RGB images from 2 classes describing the land type.\\
For Atari we used the replay buffer data released in \cite{agarwal2020optimistic} to pair the game frames with the optimal action chosen by a trained DQN agent. We combined the data from 5 different games (VideoPinball, Boxing, Breakout, StarGunner, Atlantis) to define our own domain-incremental benchmark \cite{vandeven2018} with one game per experience. Each experience contains 200k randomly sampled stacks of four consecutive game frames paired with the optimal action from the last replay buffer. In this scenario, we can treat the problem of learning the policy $\pi(a|s)$ (that predicts the action $a$ given the state $s$) as a supervised task.\\
In our Atari benchmark, the output layer is fixed since the action space is defined by Atari. However, the optimal action distribution changes across games, with some actions never being selected in some of them or their frequency drifting from one experience to the other.

\paragraph{Experimental setup.}
On each benchmark, we conducted a model selection for each calibration and CL strategy. For our RC, we kept the same configuration of the hyperparameters found during model selection on the corresponding calibration strategy (e.g., we performed model selection for TS and applied the same configuration to TS + RC).\\
We report the complete set of optimal values found by model selection in the Appendix.
On Split MNIST, we used a one-hidden-layer MLP trained with SGD. On Split CIFAR100 and EuroSAT, we used a ResNet110 and ResNet50 \cite{he2016deep}, respectively. Both models are trained with AdamW. On Atari we chose the DQN \cite{mnih15} with full Atari action space optimized with Adam. \\
% The learning rate scheduler follows a cosine annealing with warm restarts.
The reliability diagrams and the corresponding ECEs are computed from 10 equally-spaced bins. The first bin spans the $[0, 0.1]$ confidence interval, the second bin the $(0.1, 0.2]$ confidence interval and so on up until the last bin spanning the $(0.9, 1.0]$ confidence interval.\\
We used the Avalanche library \cite{carta2023} to run all the experiments. 
% At the end of training on each experience, we collected accuracy and calibration data across all test sets.
Our experiments do not use task labels. This means that at test time the model needs to distinguish between all classes learned during training.

\section{Results}

Table \ref{tab:acc} and \ref{tab:ece} report the average accuracy and ECE, respectively, on the entire data stream at the end of training. The runs are averaged over 3 random seeds.
We now highlight the main results found in our empirical evaluation.

\paragraph{Joint Training calibration.}
Calibration strategies do not hurt predictive accuracy in Joint Training, except with MS on Atari. MS also achieves the worst ECE in Joint Training in all benchmarks. These results are in line with the original MS paper \cite{guo2017}, where MS was not able to consistently train calibrated models. We will see how this behavior changes in CL setup. Interestingly, although VS performs the same type of post-processing as MS (but with a learned diagonal matrix instead of a full matrix), it shows a much better calibration in Joint Training. Again, this is aligned with the results presented in \cite{guo2017}.

\begin{figure}
\centering
\includegraphics[width=0.79\linewidth]{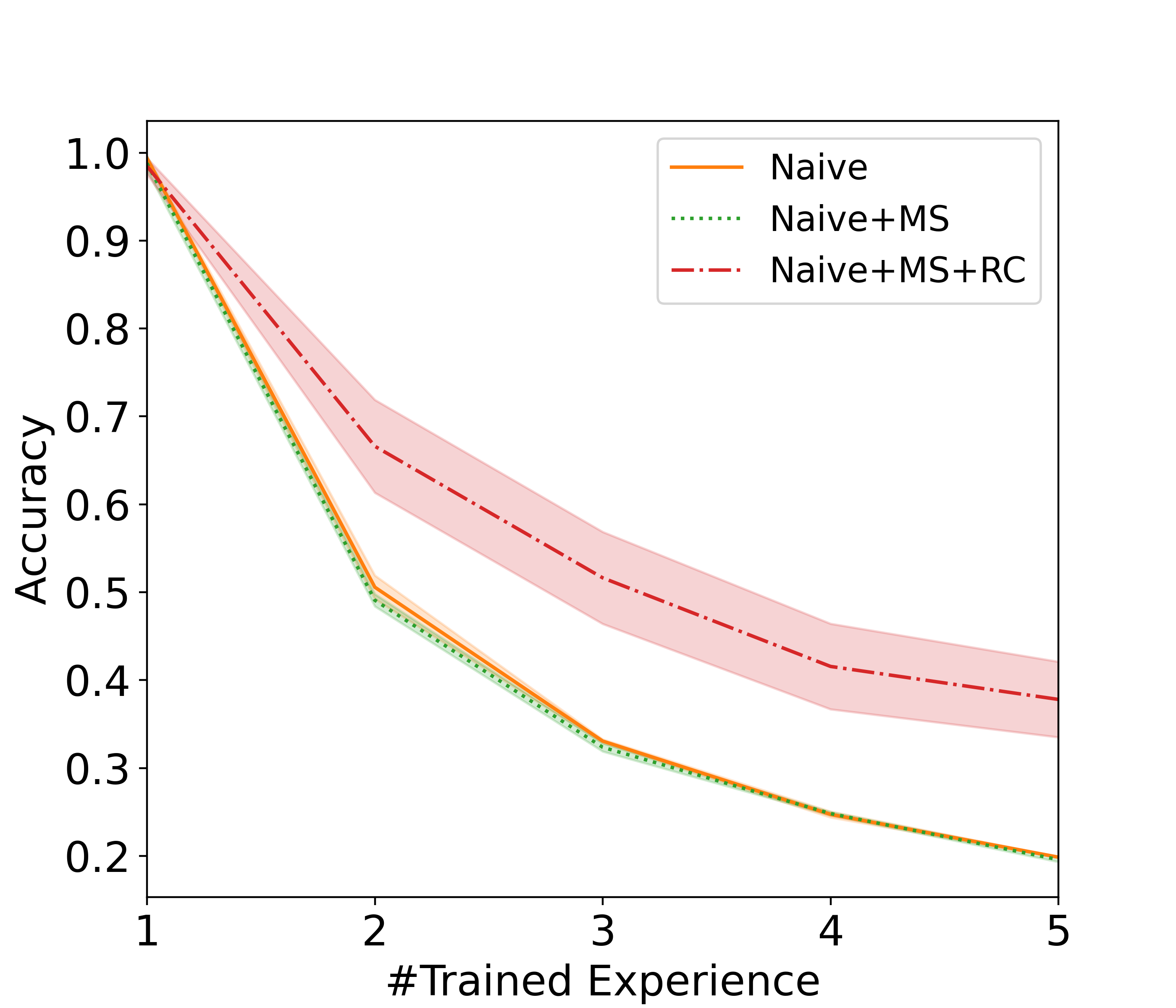}
    \caption{Accuracy of Naive on Split MNIST.}
    \label{fig:mnist-acc}
\end{figure}

\begin{figure}
\centering
\includegraphics[width=0.79\linewidth]{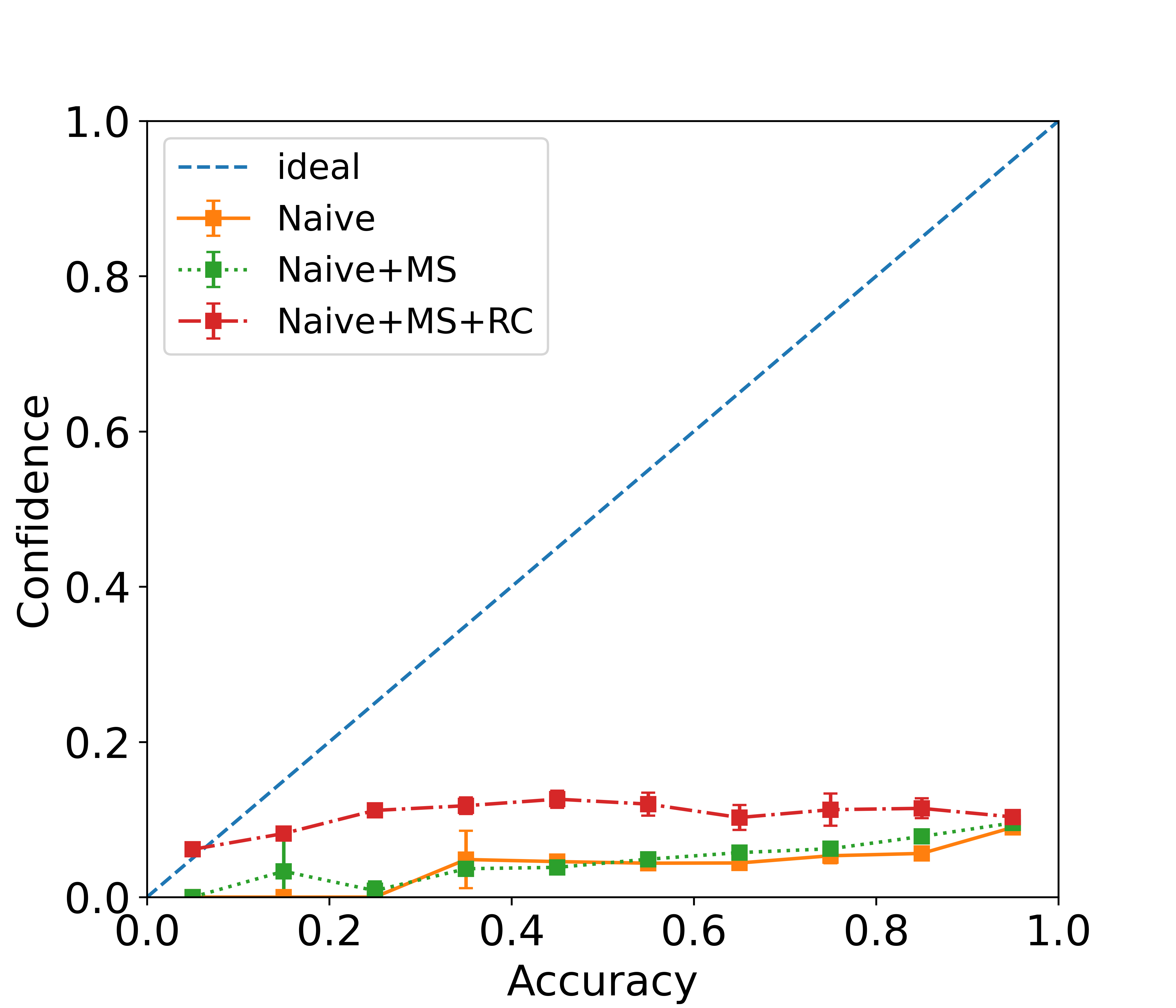}
    \caption{Calibration diagram for Naive on Split CIFAR100.}
    \label{fig:splitcifar100-naive}
\end{figure}

\paragraph{RC mitigates forgetting.}
Both the calibration and the accuracy are heavily impacted by the CL training. As expected, Naive finetuning causes catastrophic forgetting of previous knowledge on all class-incremental benchmarks. Due to its domain-incremental nature, the Atari benchmark shows a softer forgetting. The fixed output space enjoys better stability and prevents the accuracy from dropping to the level of a random classifier. When combined with MS, our RC approach is able to improve the accuracy (Figure \ref{fig:mnist-acc}) as well as the ECE on all benchmarks (Figure \ref{fig:splitcifar100-naive}). Importantly, RC is not equal to replay since the examples come from the validation set and are not used to maximize the predictive accuracy, but rather the calibration.

\begin{figure}
\centering
\includegraphics[width=0.79\linewidth]{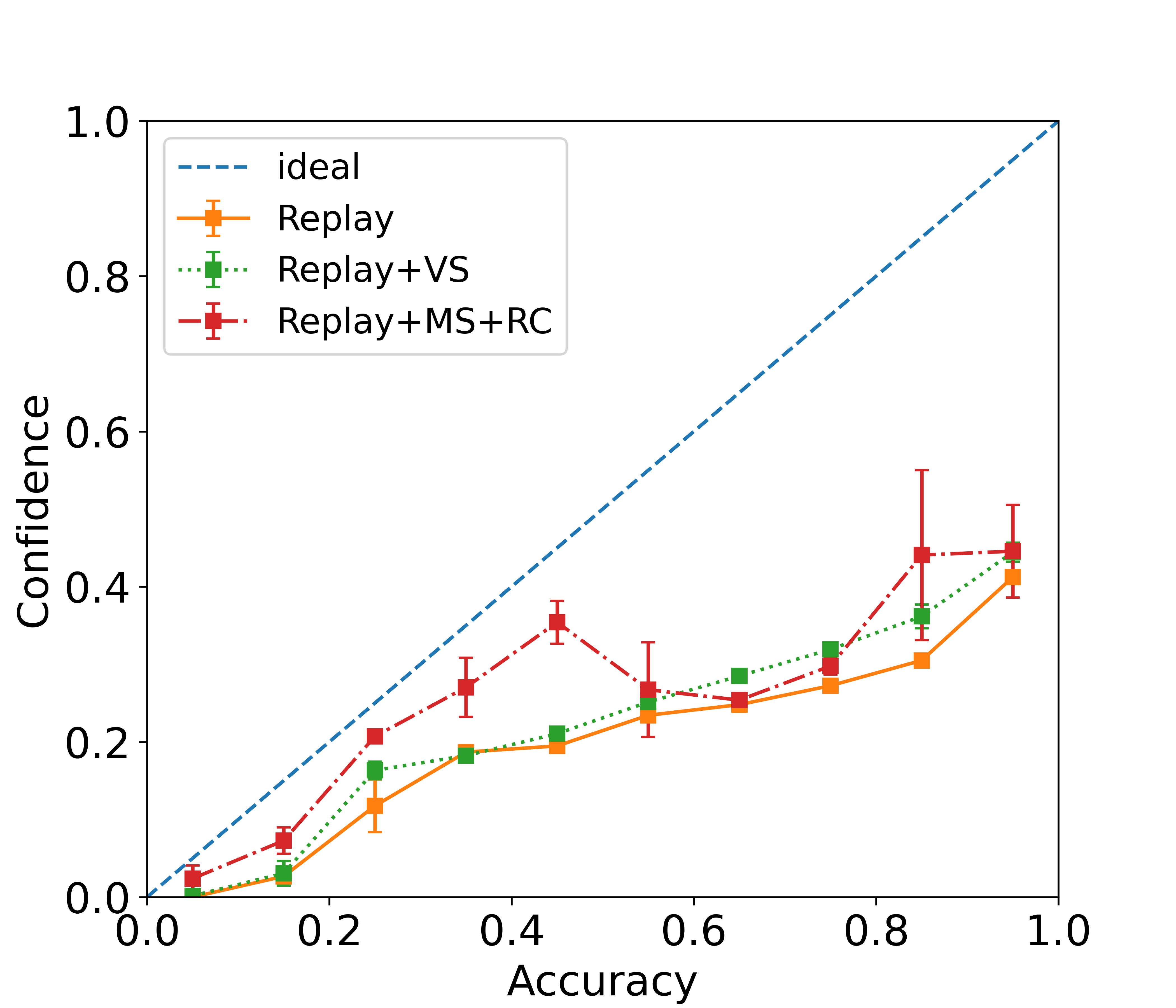}
    \caption{Calibration diagram for Replay on Atari.}
    \label{fig:atari-replay}
\end{figure}

\begin{figure}
\centering
\includegraphics[width=0.79\linewidth]{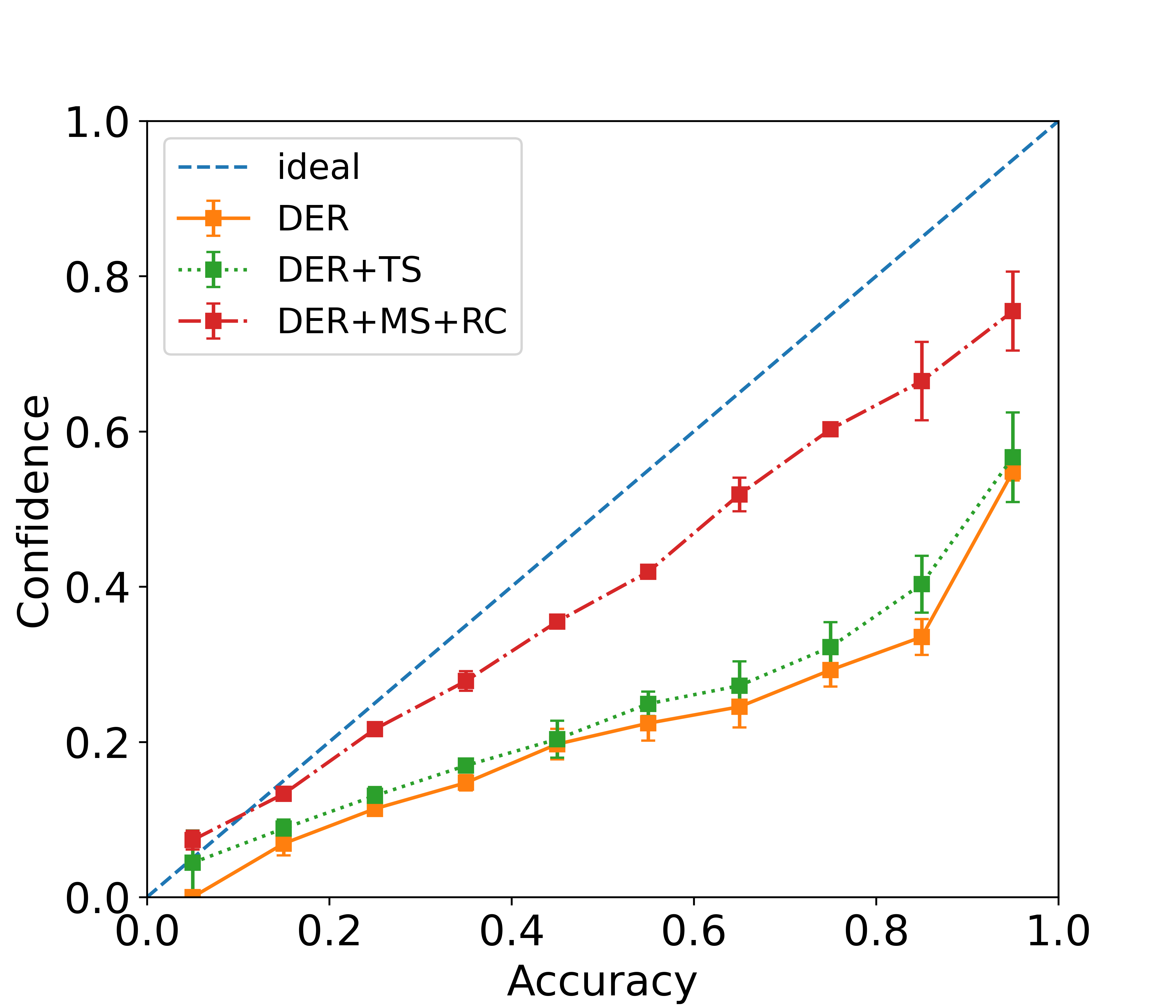}
    \caption{Calibration diagram for DER++ on Split CIFAR100.}
    \label{fig:splitcifar100-der}
\end{figure}

\paragraph{RC improves calibration.}
We found RC to be beneficial even when paired with the Replay and DER++ strategies (Figure \ref{fig:atari-replay} and Figure \ref{fig:splitcifar100-der}, respectively). Although there is no unique post-processing strategy that consistently performs better than the others, our RC always improves calibration on all 4 benchmarks, often by large margins. For example, on EuroSAT, post-processing strategies alone achieve an ECE between 12\% (best case) and 15\% (worse). After applying RC, we are able to reduce the gap to 4\% and 9\%, respectively.\\
In some cases, a calibrated model results in a decrease in accuracy. For example, TS on Split CIFAR100 drops the average accuracy from $40\%$ with Replay to $23\%$ with TS and $15\%$ with TS+RC. Still, in terms of calibration TS outperforms the other approaches. This trade-off needs to be carefully considered: whether to prefer a less accurate, but calibrated model, or vice versa a more accurate but less calibrated model. However, it is also important to note that calibration does not necessarily causes a drop in accuracy. 
For example, Replay with VS+RC on EuroSAT results in an improvement in both accuracy and ECE.
% Interestingly, these runs always showed a large standard deviation. A similar phenomenon happens with MS. In both cases, the intervention on the output layer (by either changing the softmax temperature or by adding a completely new output layer) result in a more unstable performance. We did not observe the same behavior for the self-calibration strategy HR, which remains stable over the course of training.

\paragraph{Calibration techniques boost DER++.}
DER++ is one of the most interesting strategies in terms of calibration. DER++ achieves the best calibration on the very challenging Split CIFAR100 and Atari. The original paper \cite{buzzega2020} also showed the effectiveness of DER++ in calibrating CL models. However, the paper did not leverage any calibration methods and did not report precise calibration values. In our experiments, we show how DER++ is indeed very effective, but we also point out how its calibration ability can easily be improved when coupled with calibration strategies. In particular, our RC strategy outperforms all other combinations when coupled with MS (Figure \ref{fig:splitcifar100-der}), without a substantial decrease in accuracy.\\

Calibration of CL models remains less effective than that of Joint Training models. However, we showed how CL strategies and calibration strategies can operate together, resulting in better calibrated models. Even a state-of-the-art strategy like DER++ enjoys clear improvements in calibration with post-processing techniques. \\
Our results apply to a diverse set of CL benchmarks, some of which are especially promising in terms of real-world applications (EuroSAT) and generalization to other kinds of problems beyond pure pattern recognition (Atari).

\section{Conclusion and Future Work}
We start from the assumption that perfect predictive models do not exist. CL models inevitably make mistakes. Instead of only pursuing a better predictive model, we argue that it is equally relevant to design robust models that can be trusted. Calibration allows the model itself to learn a meaningful notion of confidence about its predictions. In particular, the confidence expresses the expected average accuracy on that kind of examples. Calibrated models can operate more autonomously than uncalibrated ones since they can detect when they are likely to make a mistake and call for external help (e.g., a human). \\

We provided the first empirical evaluation on continual calibration and we show how, on one side, CL models are not naturally calibrated and that, on the other side, post-processing calibration and self-calibration are effective when combined with CL strategies. Our Replayed Calibration improved the performance of post-processing calibration methods across different calibration and CL techniques. We hope our work can increase the attention towards continual calibration, and we highlight some promising research directions. \\
There are only a few self-calibration techniques currently available for multi-class classification with neural networks \cite{pereyra2017a}. However, self-calibration techniques are inherently compatible with a CL setup, since they operate online during training, without requiring a separate calibration phase. Efforts in designing new self-calibration techniques could directly benefit their CL application. \\
Calibration mostly considers supervised classification tasks. It is not entirely clear how to frame calibration in other types of tasks, like Reinforcement Learning. A recent work on the topic shed some light on this very challenging research direction in a model-based setting \cite{malik2019}. Being non-stationary by design, discoveries in calibration for reinforcement learning would likely have a strong impact on CL as well.\\
Finally, our empirical evaluation considered several CL benchmarks, including real-world datasets like EuroSAT and action classification from Atari. We are planning to extend our experiments by also including Natural Language Processing benchmarks. We look forward to future works tackling the continual calibration challenge.
% Since next-token prediction is a classification task with a fixed output space (the vocabulary), we can directly apply the calibration techniques considered in this work. 

\section*{Acknowledgements}
Work supported by EU EIC project EMERGE (Grant No. 101070918) and by PNRR- M4C2- Investimento 1.3, Partenariato Esteso PE00000013- "FAIR- Future Artificial Intelligence Research"- Spoke 1 "Human-centered AI", funded by the European Commission under the NextGeneration EU programme.

{
    \small
    \bibliographystyle{ieeenat_fullname}
    \bibliography{mainbib}
}

\clearpage
\setcounter{page}{1}
\maketitlesupplementary

\section{Experimental setup}
\label{sec:expsetup}
%For our task specific experiments both in CL scenario and No-CL scenario, the following parameters are used.
We report the optimal hyperparameters for all the benchmarks, different CL strategies and calibration techniques. Table \ref{tab:clsetup} provides the best hyper-parameters for the CL training. Table \ref{tab:calibrationsetup} provides the same information about calibration techniques.\\
In Split MNIST and Atari we use SGD and Adam respectively with default values. In Split CIFAR100 and EuroSAT the chosen optimizer is AdamW (\textit{weight\_decay} $= 0.0005$) and we adopt as learning rate scheduler Cosine Annealing with Warm Restarts (\textit{first restart iteration} $= 5$, \textit{minimum lr} $= 0.00001$).  
%In Split MNIST the model is a MLP with one hidden layer of size 512, in SplitCIFAR100 and EuroSAT we use ResNet \cite{he2016deep} with 50 and 110 depth respectively, lastly in Atari we follow the DQN model from \cite{mnih15}. 
For all the post-processing calibration techniques we fixed the number of training iterations to 100.\\

\begin{table}
    \centering
    \caption{Best hyper-parameters for the continual training phase. }
    \label{tab:clsetup}
    \small
    \begin{tabular}{c|cccc}
    \toprule
    & \textit{S. MNIST}& \textit{S. CIFAR100}& \textit{EuroSAT}& \textit{Atari} \\
    \midrule
    % \hline & \multicolumn{4}{c}{General}\\
    % \hline
    lr & 1e-3 & 1e-2 & 1e-3 & 5e-4 \\
    mb size & $32$ & $128$ & $128$ & $256$ \\
    epochs & $20$ & $50$ & $20$ & $100$ \\
    validation split & $0.2$ & $0.2$ & $0.1$ & $0.2$ \\
    patience & $-$ & $-$& $-$& $10$ \\
    memory size & $2000$ & $4000$ & $2000$& $4000$ \\
    \midrule
    \multicolumn{5}{c}{DER++}\\
    \midrule
    $\alpha$ & $0.3$ & $0.2$ & $0.1$ & $0.5$ \\
    $\beta$ & $0.8$ & $0.8$ & $0.5$ & $0.5$ \\
    \bottomrule
    \end{tabular}
\end{table}

\begin{table}
    \centering
    \caption{Best hyperparameters for the calibration approaches.}
    \label{tab:calibrationsetup}
    \begin{tabular}{c|cccc}
    \toprule
    & \textit{S. MNIST}& \textit{S. CIFAR100}& \textit{EuroSAT}& \textit{Atari} \\
    \midrule
    \multicolumn{5}{c}{Joint}\\
    \midrule
    ST - $\lambda$ & $0.0075$ & $0.025$ & $0.0075$ & $0.0075$ \\
    TS - lr & $0.01$& $0.01$& $0.01$& $0.001$ \\
    VS - lr & $0.01$& $0.01$& $0.01$& $0.001$ \\
    MS - lr & $0.01$& $0.01$& $0.01$& $0.01$ \\
    \midrule
    \multicolumn{5}{c}{DER++}\\
    \midrule
    HR - $\lambda$ & $0.005$ & $0.025$ & $0.025$ & $0.025$ \\
    TS - lr & $0.01$& $0.01$& $0.01$& $0.01$ \\
    VS - lr & $0.01$& $0.01$& $0.01$& $0.01$ \\
    MS - lr & $0.01$& $0.01$& $0.01$& $0.01$ \\
    \midrule
    \multicolumn{5}{c}{Replay}\\
    \midrule
    HR - $\lambda$ & $0.025$ & $0.025$ & $0.025$ & $0.0025$ \\
    TS - lr & $0.01$& $0.01$& $0.01$& $0.01$ \\
    VS - lr & $0.01$& $0.01$& $0.01$& $0.001$ \\
    MS - lr & $0.01$& $0.01$& $0.01$& $0.001$ \\
    \midrule
    \multicolumn{5}{c}{Naive}\\
    \midrule
    HR - $\lambda$ & $0.075$ & $0.1$ & $0.0075$ & $0.005$ \\
    TS - lr & $0.01$& $0.01$& $0.01$& $0.001$ \\
    VS - lr & $0.01$& $0.01$& $0.01$& $0.001$ \\
    MS - lr & $0.01$& $0.01$& $0.01$& $0.001$ \\
    \bottomrule
    \end{tabular}
\end{table}

\section{DER++ implementation}
We used the DER++ version present in Avalanche \cite{carta2023}. Since the experimental setup and the details of the implementation may differ between the original version \cite{buzzega2020} and the Avalanche version, we ran some experiments to compare the performance. Table \ref{tab:der} shows that the average test accuracy on Split CIFAR100 and Split TinyImageNet obtained at the end of training does not change. \\
We used this sanity-check to ensure that the calibration performance of DER++ does not depend on a custom DER++ version. 

\section{Sensitivity of MS to changes in the learning rate}
\begin{figure}
\centering
\includegraphics[width=\linewidth]{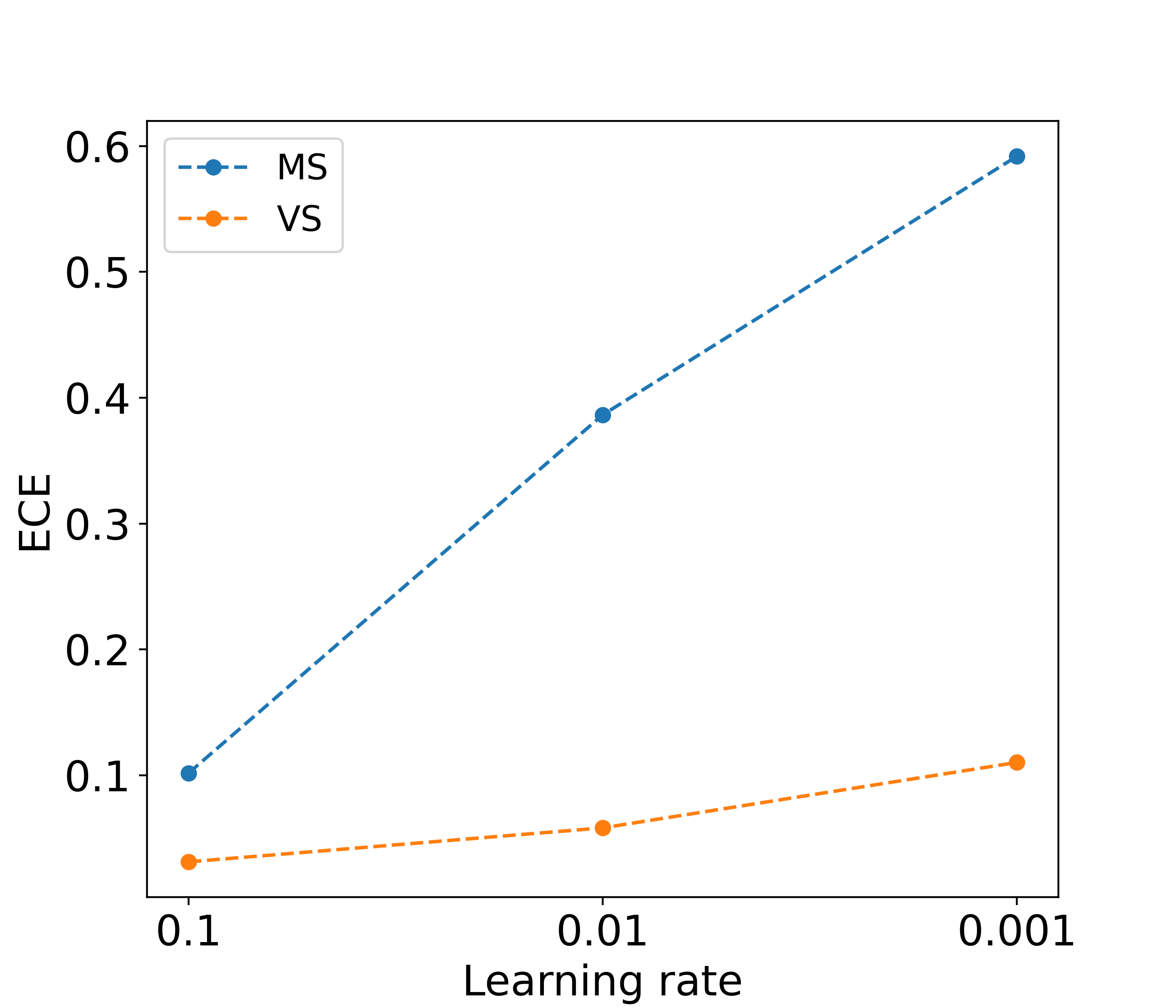}
    \caption{Sensitivity of Joint Training and MS to the learning rate. The dataset is Split CIFAR 100. The ECE does not change much for VS, while MS shows a large sensitivity on the chosen learning rate.}
    \label{fig:sensitivity}
\end{figure}

Figure \ref{fig:sensitivity} shows that calibration with MS on Joint Training is very sensitive to the choice of the learning rate. The ECE jumps from 10\% to 60\%, depending on the chosen learning rate.

\section{Decision between retaining or discarding the wrapped model after calibration}
\begin{figure}
\centering
\includegraphics[width=\linewidth]
{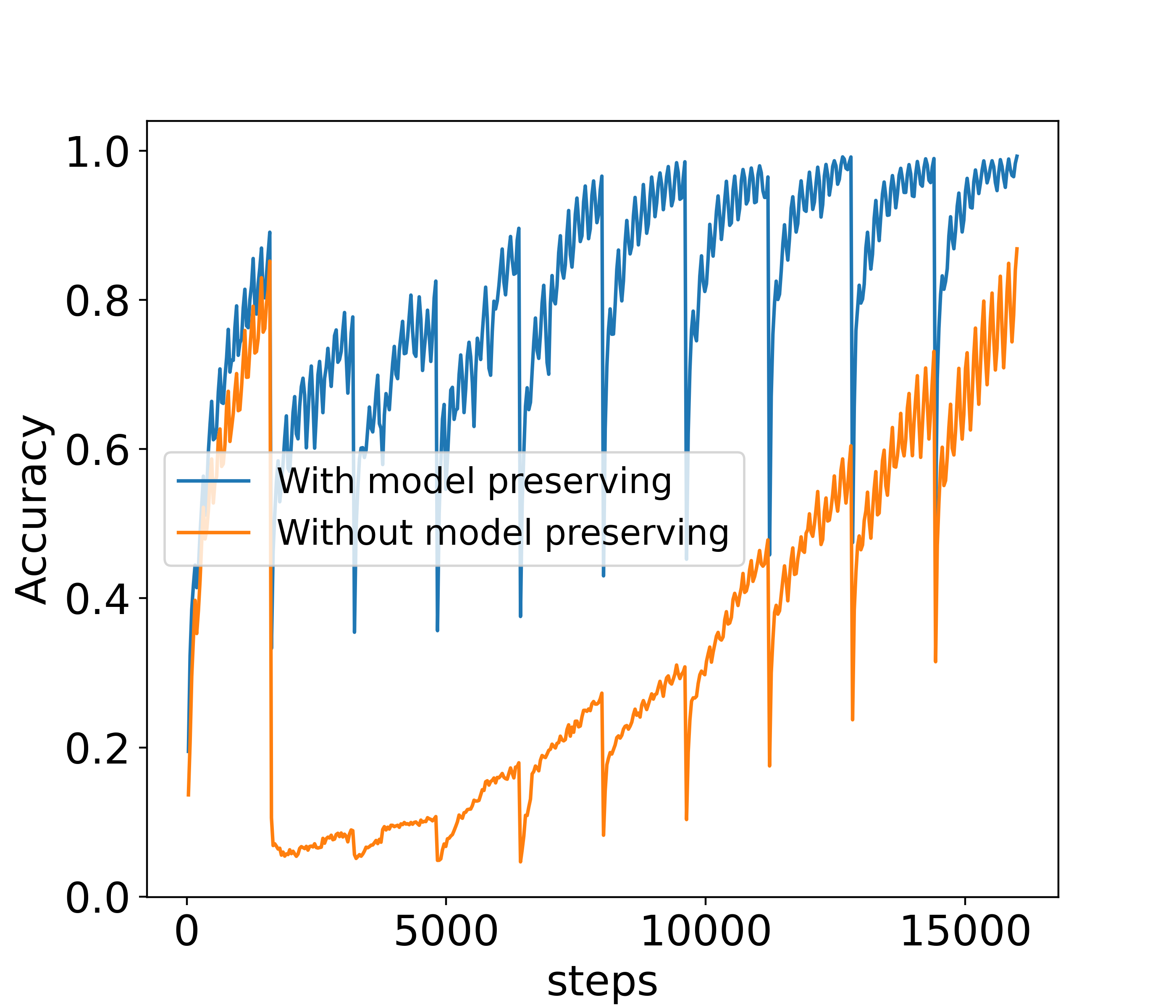}
    \caption{Comparison between re-training and discarding the wrapped model after the calibration phase.}
    \label{fig:comparison_retaining_discarding_mode}
\end{figure}

In the post-processing Calibration method, we adjust the softmax temperature after the output layer or introduce additional linear projection during the calibration phase through temperature scaling or vector/matrix scaling. In CL scenarios, we encounter a sequence of experiences, where each experience concludes with a calibration phase following training. This alternation between training and calibration phases presents the option to either retain the wrapped model after the calibration phase or discard it for each training phase, utilizing it exclusively during calibration. We conduct experiments to explore both approaches. Figure \ref{fig:comparison_retaining_discarding_mode} demonstrates a scenario involving training the DER model with Adamw + replayed matrix scaling calibration on the Cifar100 dataset. Here, we compare the accuracy between retaining and discarding the wrapped calibration model after the calibration phase. Notably, discarding the wrapped model results in complete forgetting after the first experience, necessitating the model to essentially "re-learn" as depicted in the figure. Conversely, retaining the wrapped model showcases a more stable learning curve, yielding higher accuracy and lower ECE. Based on these experimental observations, we choose to preserve the wrapped model after each calibration phase for all post-processing calibration experiments.
\color{black}

\begin{table}[htbp]
  \centering
  \caption{Comparison between the published results from DER++ and results obtained with our implementation of DER++ on Split CIFAR100 and Split Tiny ImageNet. We successfully replicate the results from the original papers.}
  \label{tab:der}
  \small
  \begin{tabular}{c|cc}
  \toprule
  \textbf{Accuracy (\%)} & \textit{\textbf{S. CIFAR100}} \cite{boschini2022a} & \textbf{\textit{S. TinyImageNet}} \cite{buzzega2020} \\
  \midrule
  Joint & $70.44$ & $59.99 \pm 0.19$ \\
  DER++ & $53.63$ & $10.96 \pm 1.17$ \\
  Replay & $38.58$ & $8.49 \pm 0.16$ \\
  Naive & $9.43$ & $7.92 \pm 0.26$ \\
  \hline
  \hline
  Joint (\textit{ours}) & $69.00 \pm 4.96$ & $62.00 \pm 0.52$ \\
  DER++ (\textit{ours}) & $51.91 \pm 0.93$ & $12.83 \pm 0.30$ \\
  Replay (\textit{ours}) & $40.47 \pm 0.95$ & $10.10 \pm 0.28$ \\
  Naive (\textit{ours}) & $9.07 \pm 0.10$ & $7.52 \pm 0.04$ \\
  \bottomrule
  \end{tabular}
\end{table}

\section{Reliability diagrams}
We report the complete set of reliability diagrams for each benchmark and strategy.
% Plots follow same order of the tables

\begin{figure}
\centering
\includegraphics[width=\linewidth]{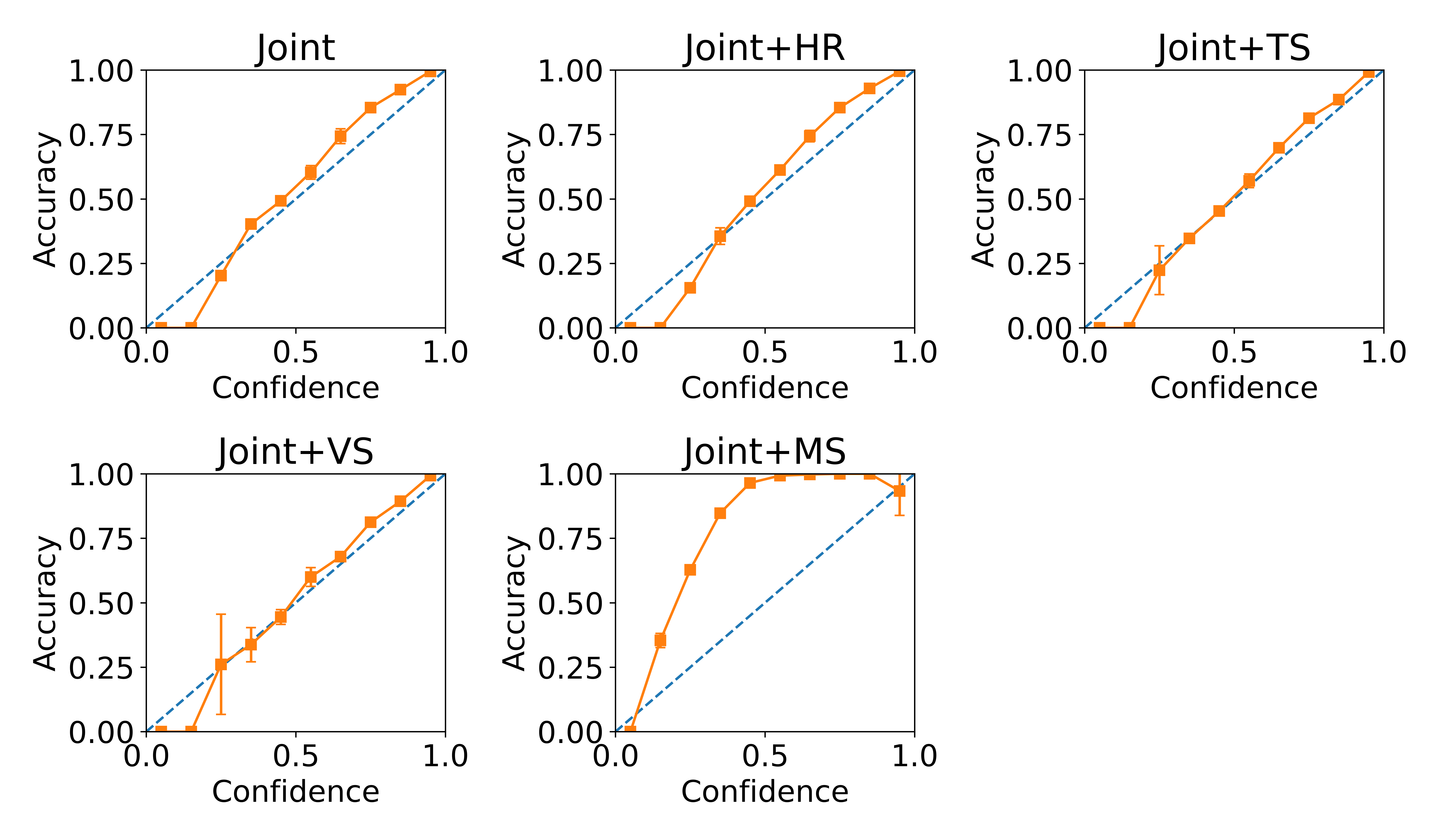}
    \caption{Reliability diagrams for Joint on Split MNIST}
    \label{fig:splitmnist-all-joint}
\end{figure}

\begin{figure}
\centering
\includegraphics[width=\linewidth]{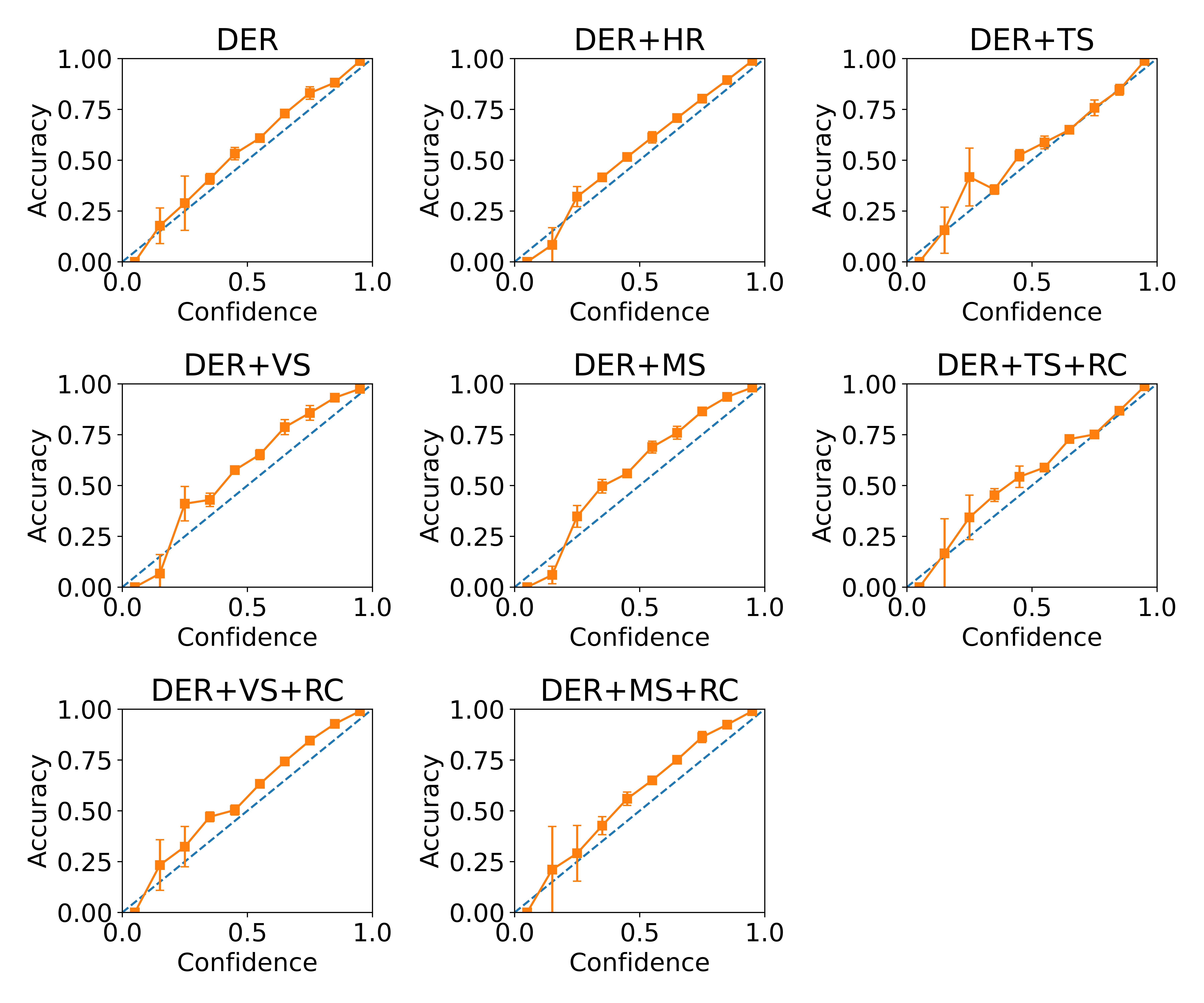}
    \caption{Reliability diagrams for DER++ on Split MNIST}
    \label{fig:splitmnist-all-der}
\end{figure}

\begin{figure}
\centering
\includegraphics[width=\linewidth]{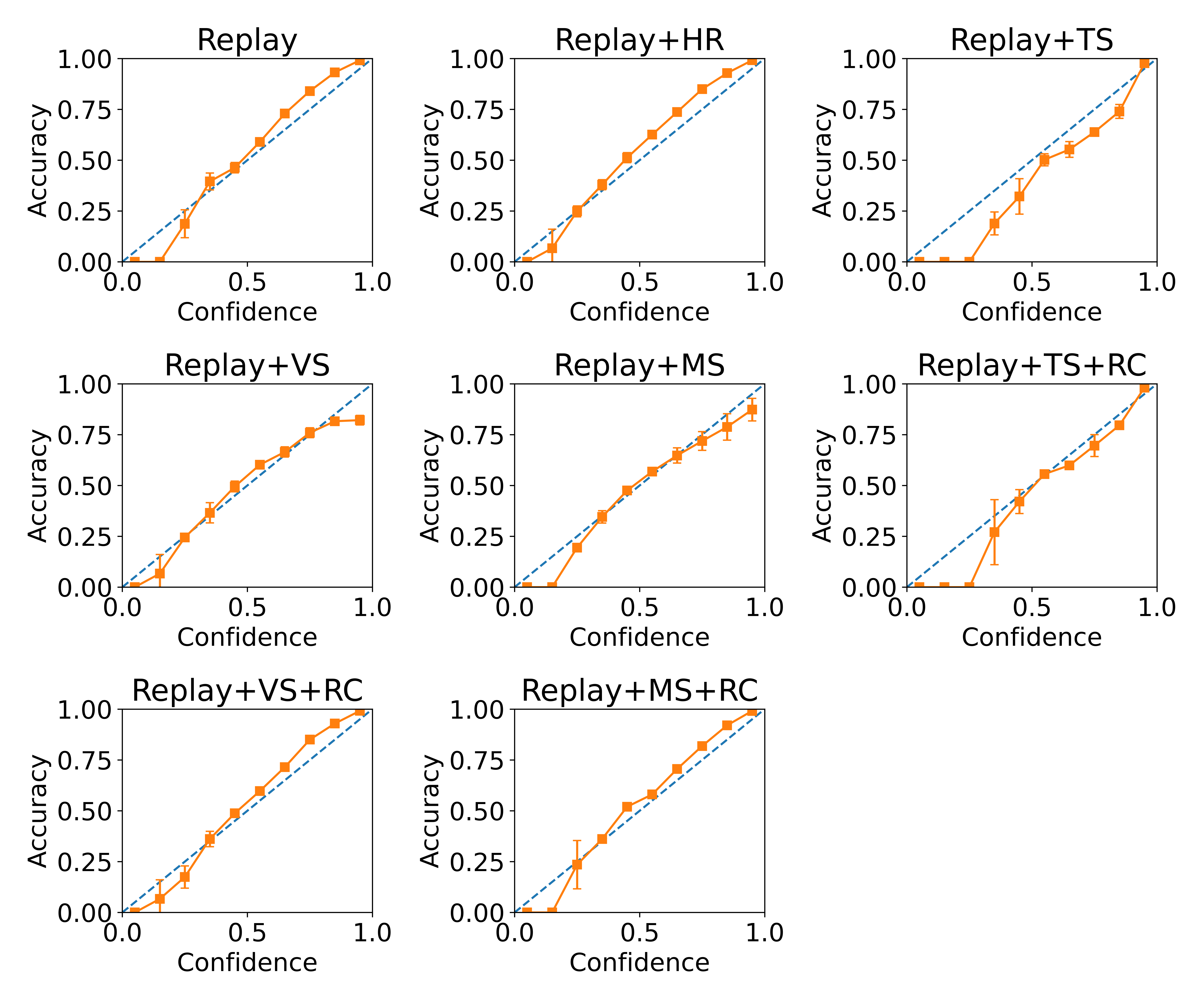}
    \caption{Reliability diagrams for Replay on Split MNIST}
    \label{fig:splitmnist-all-replay}
\end{figure}

\begin{figure}
\centering
\includegraphics[width=\linewidth]{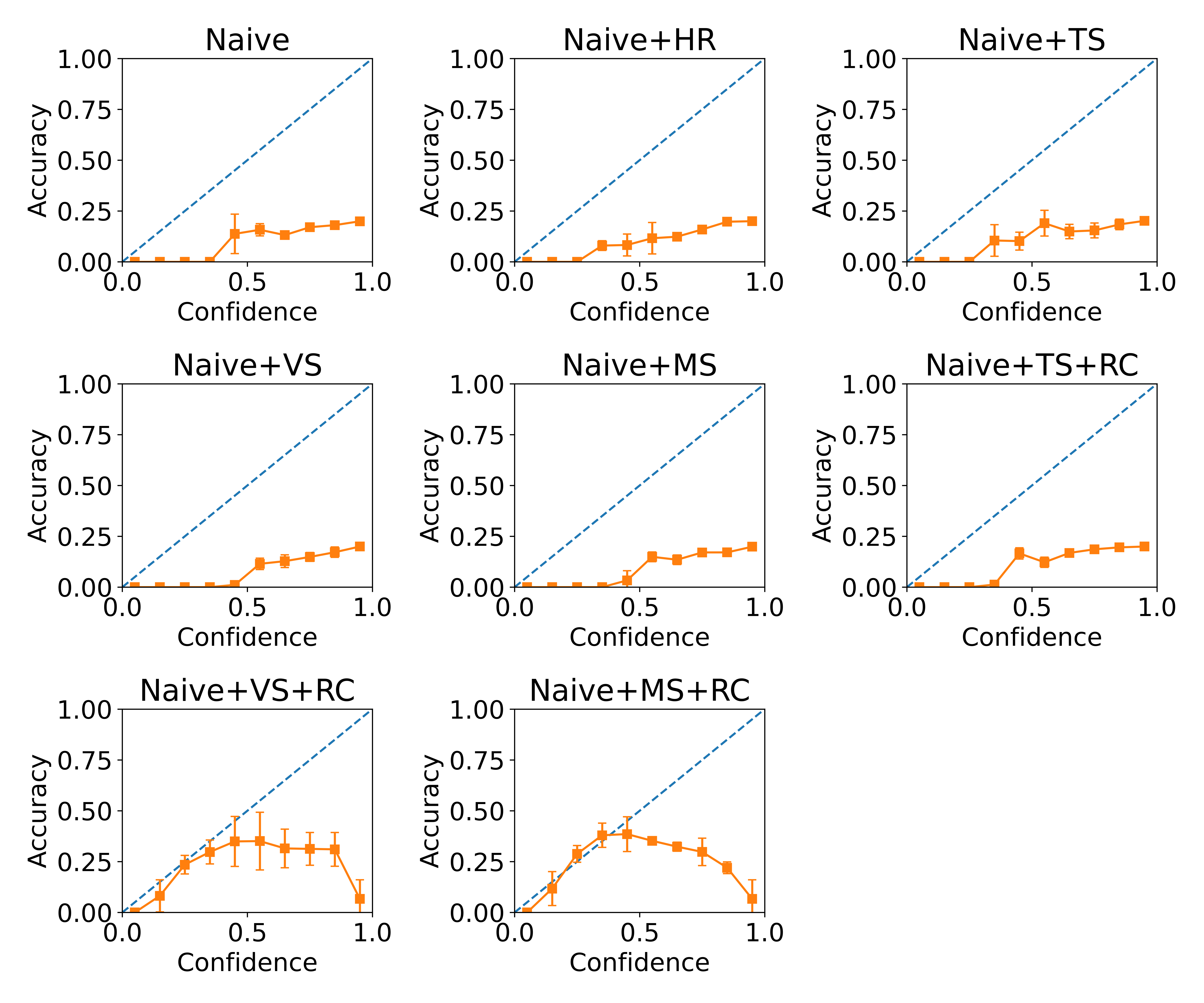}
    \caption{Reliability diagrams for Naive on Split MNIST}
    \label{fig:atari-smnist-naive}
\end{figure}

\begin{figure}
\centering
\includegraphics[width=\linewidth]{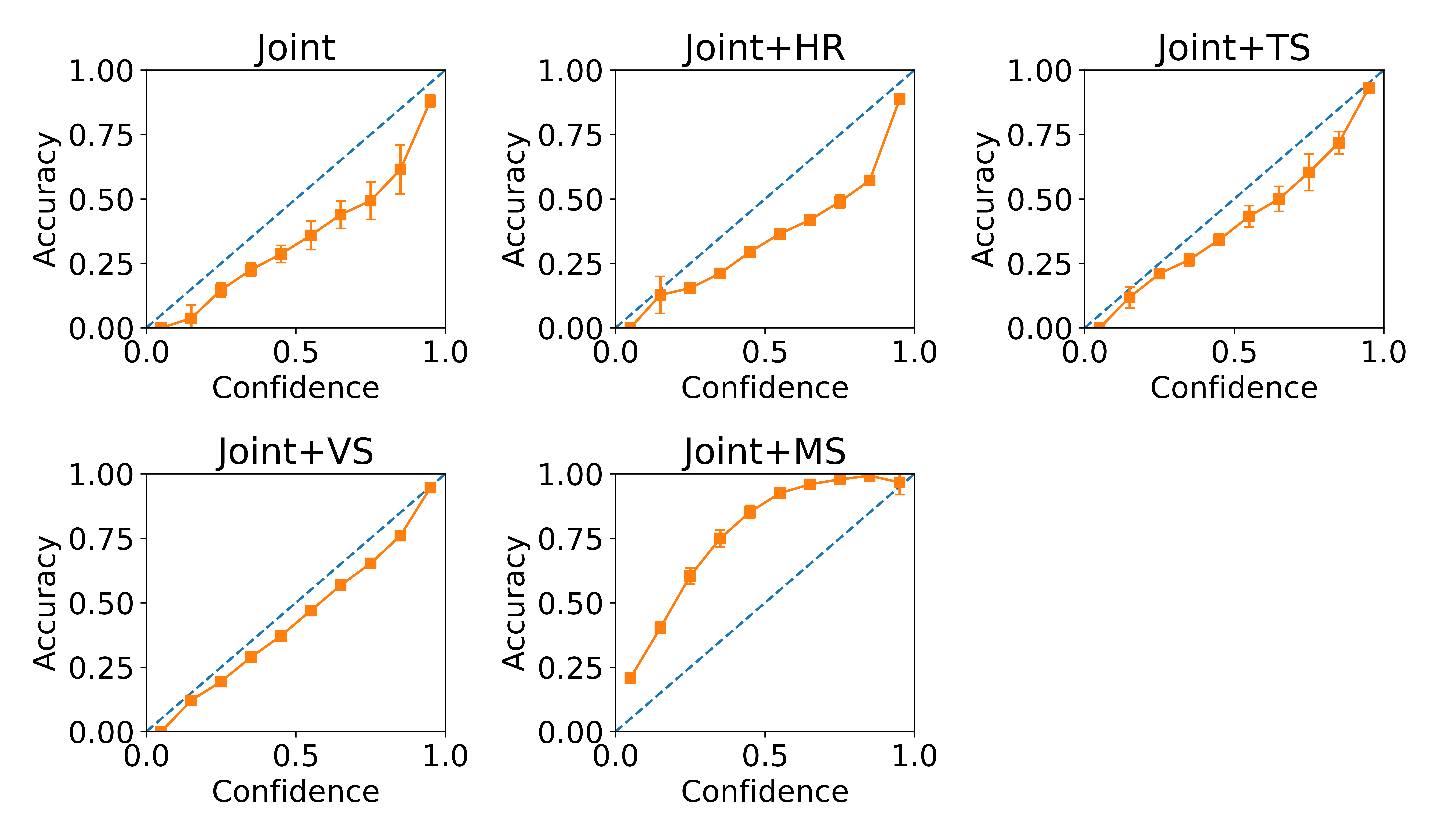}
    \caption{Reliability diagrams for Joint on Split CIFAR100}
    \label{fig:splitcifar100-all-joint}
\end{figure}

\begin{figure}
\centering
\includegraphics[width=\linewidth]{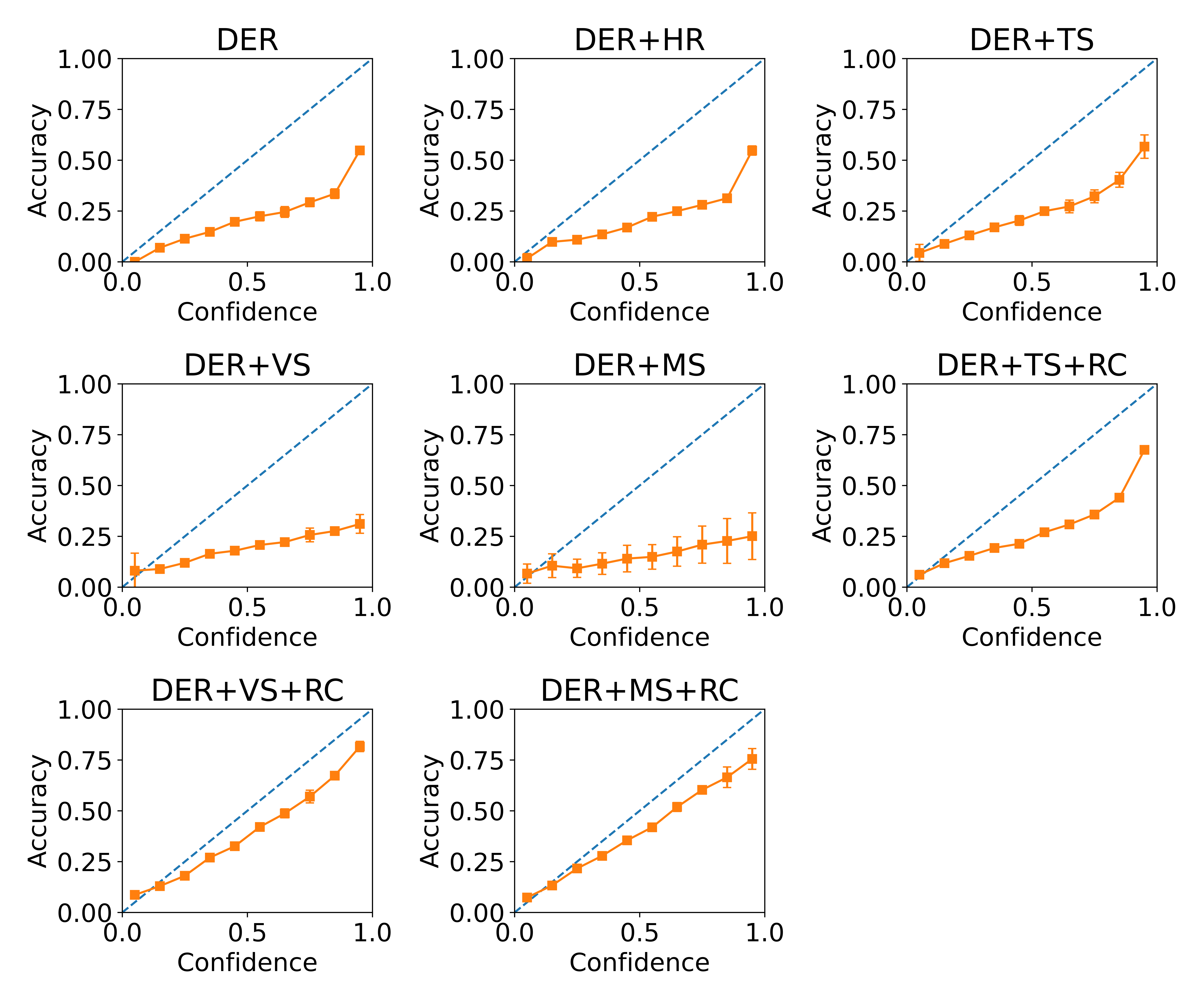}
    \caption{Reliability diagrams for DER++ on Split CIFAR100}
    \label{fig:splitcifar100-all-der}
\end{figure}

\begin{figure}
\centering
\includegraphics[width=\linewidth]{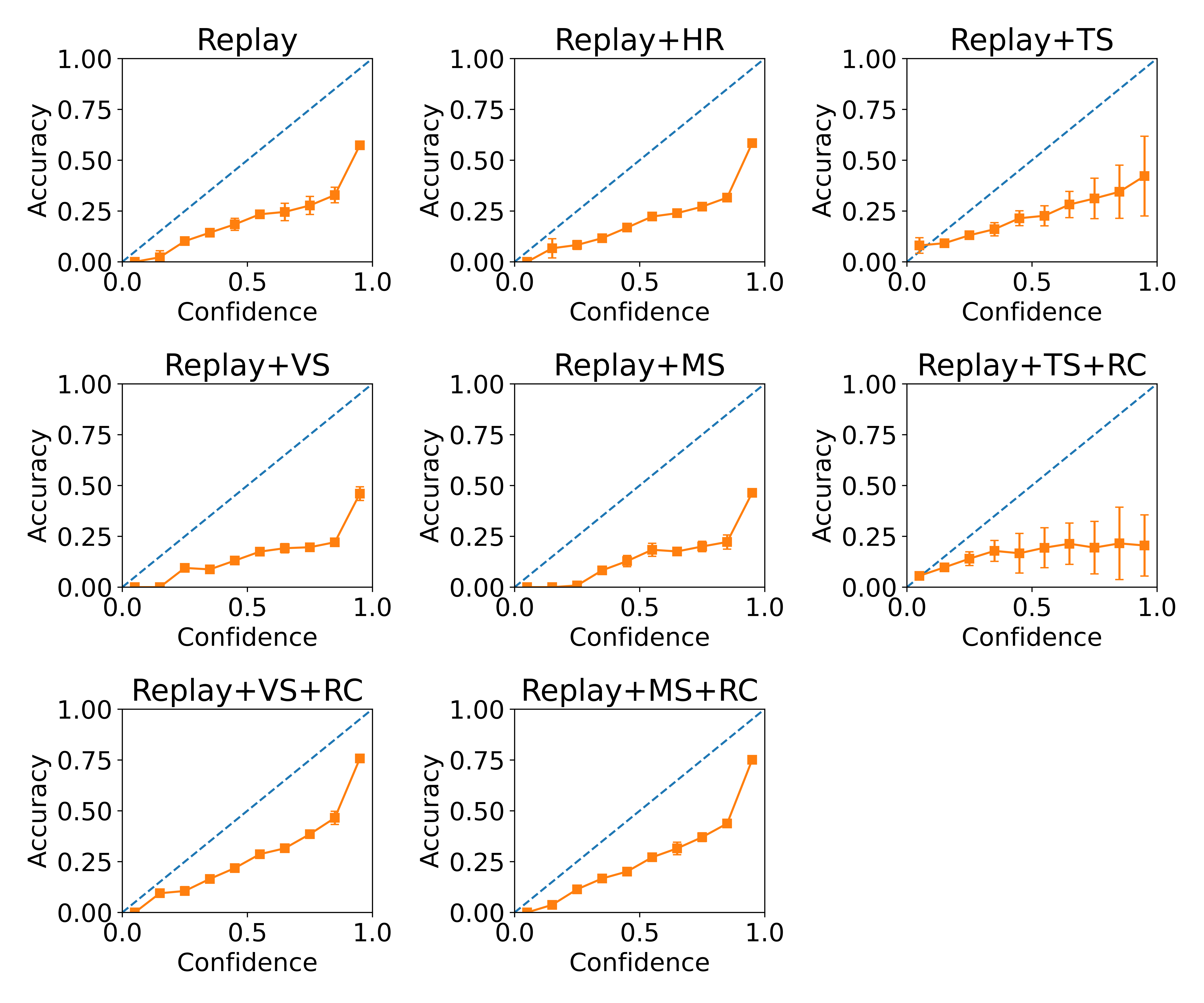}
    \caption{Reliability diagrams for Replay on Split CIFAR100}
    \label{fig:splitcifar100-all-replay}
\end{figure}

\begin{figure}
\centering
\includegraphics[width=\linewidth]{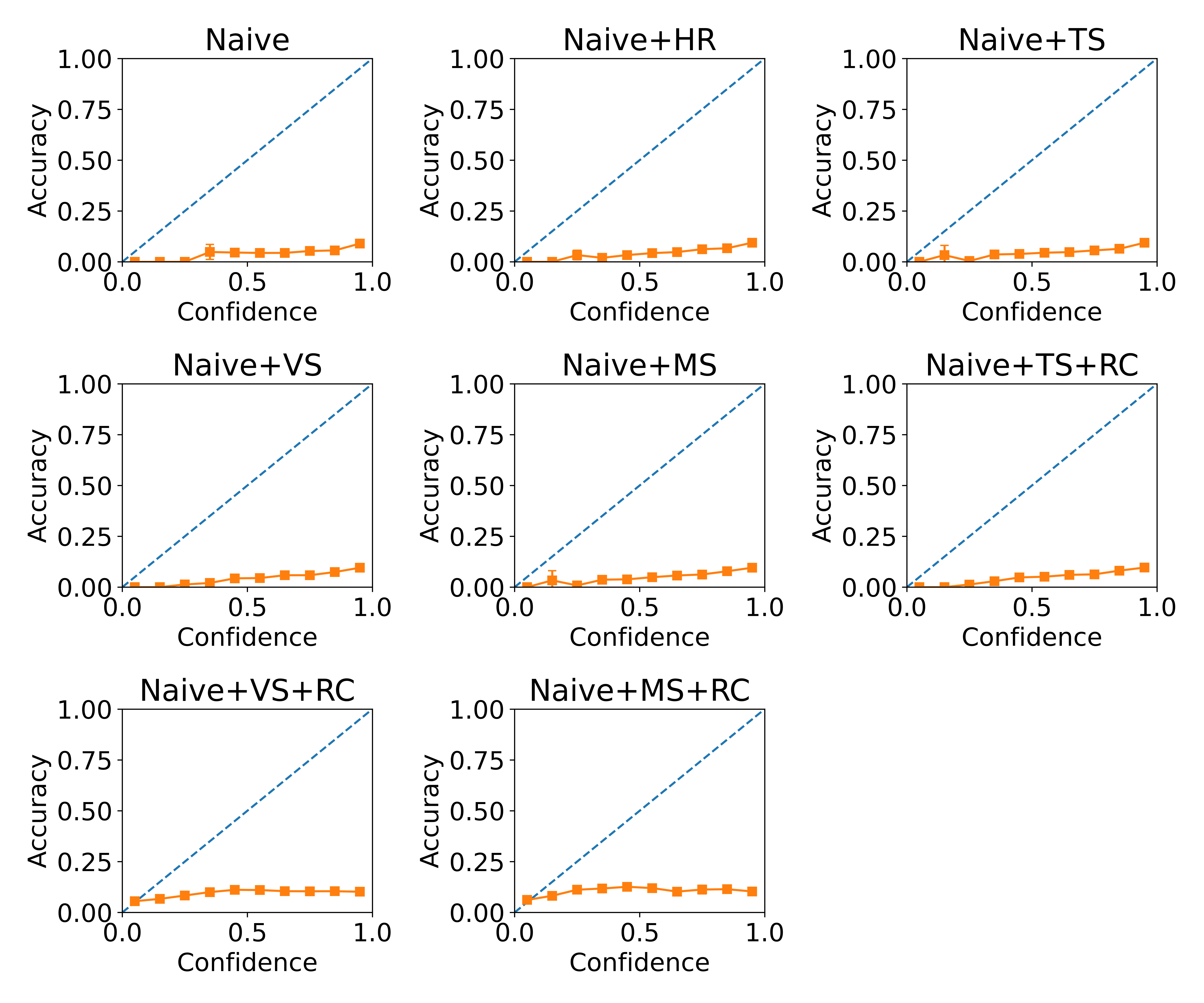}
    \caption{Reliability diagrams for Naive on Split CIFAR100}
    \label{fig:atari-scifar-naive}
\end{figure}

\begin{figure}
\centering
\includegraphics[width=\linewidth]{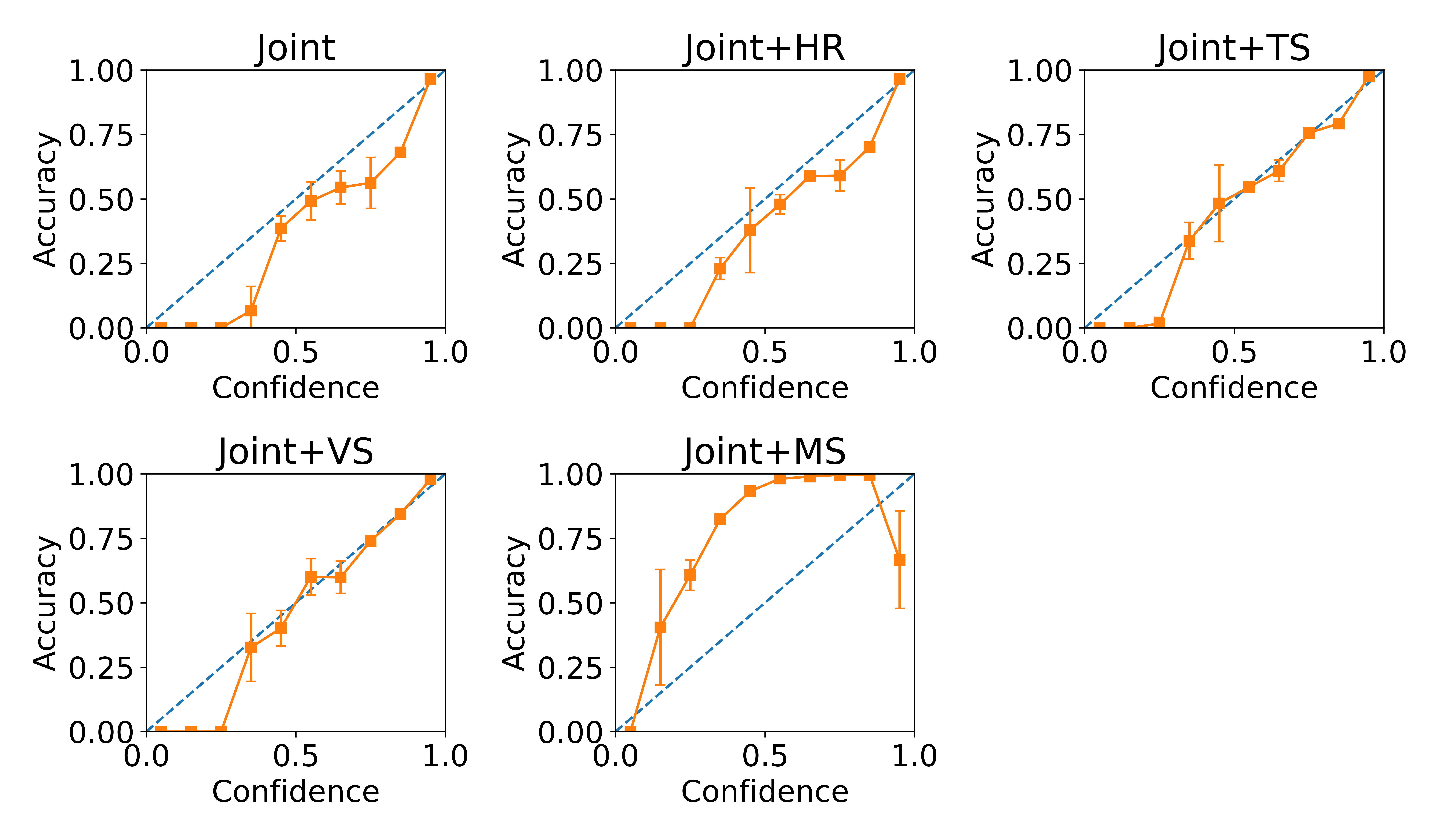}
    \caption{Reliability diagrams for Joint on EuroSAT}
    \label{fig:eurosat-all-joint}
\end{figure}

\begin{figure}
\centering
\includegraphics[width=\linewidth]{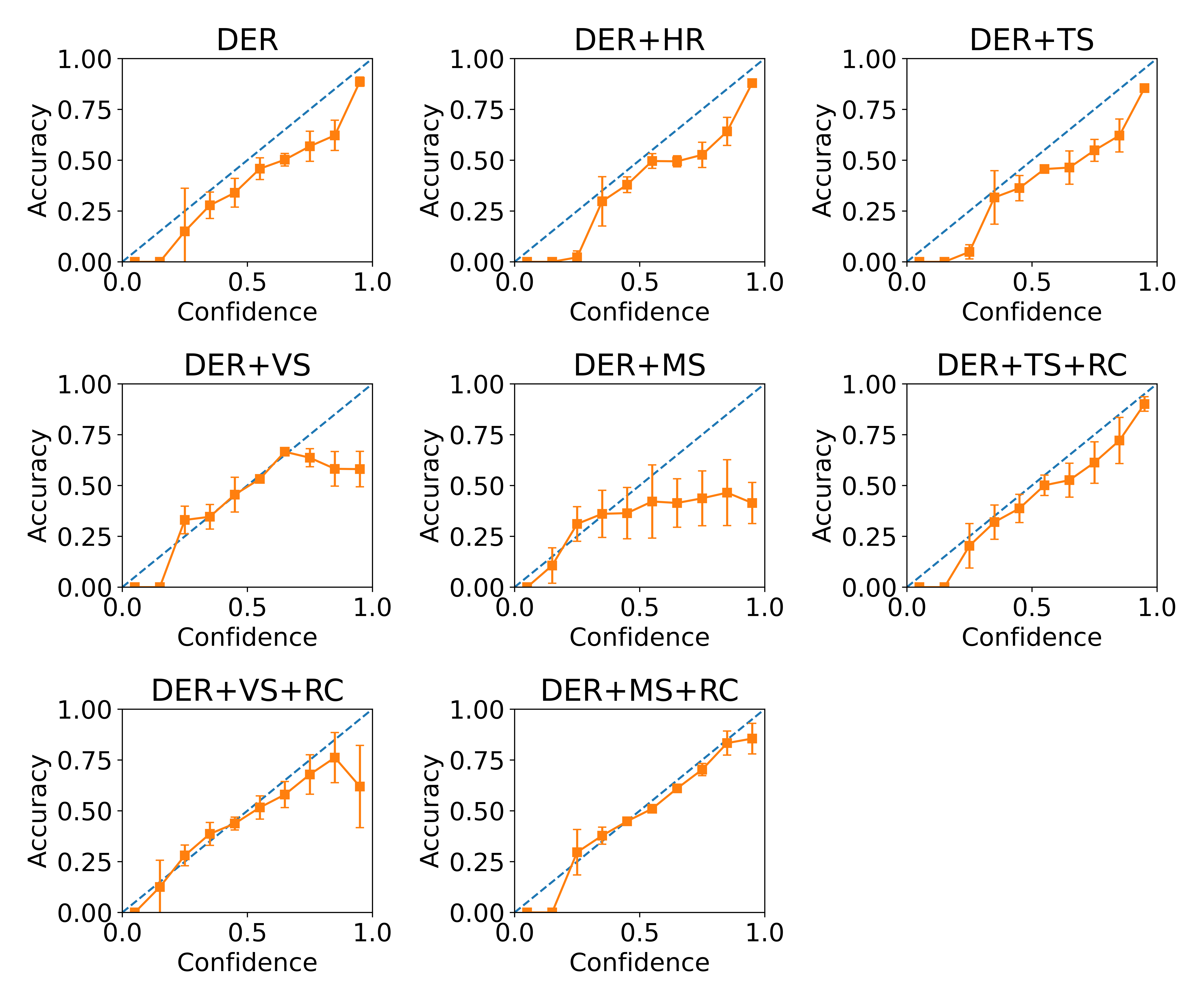}
    \caption{Reliability diagrams for DER++ on EuroSAT}
    \label{fig:eurosat-all-der}
\end{figure}

\begin{figure}
\centering
\includegraphics[width=\linewidth]{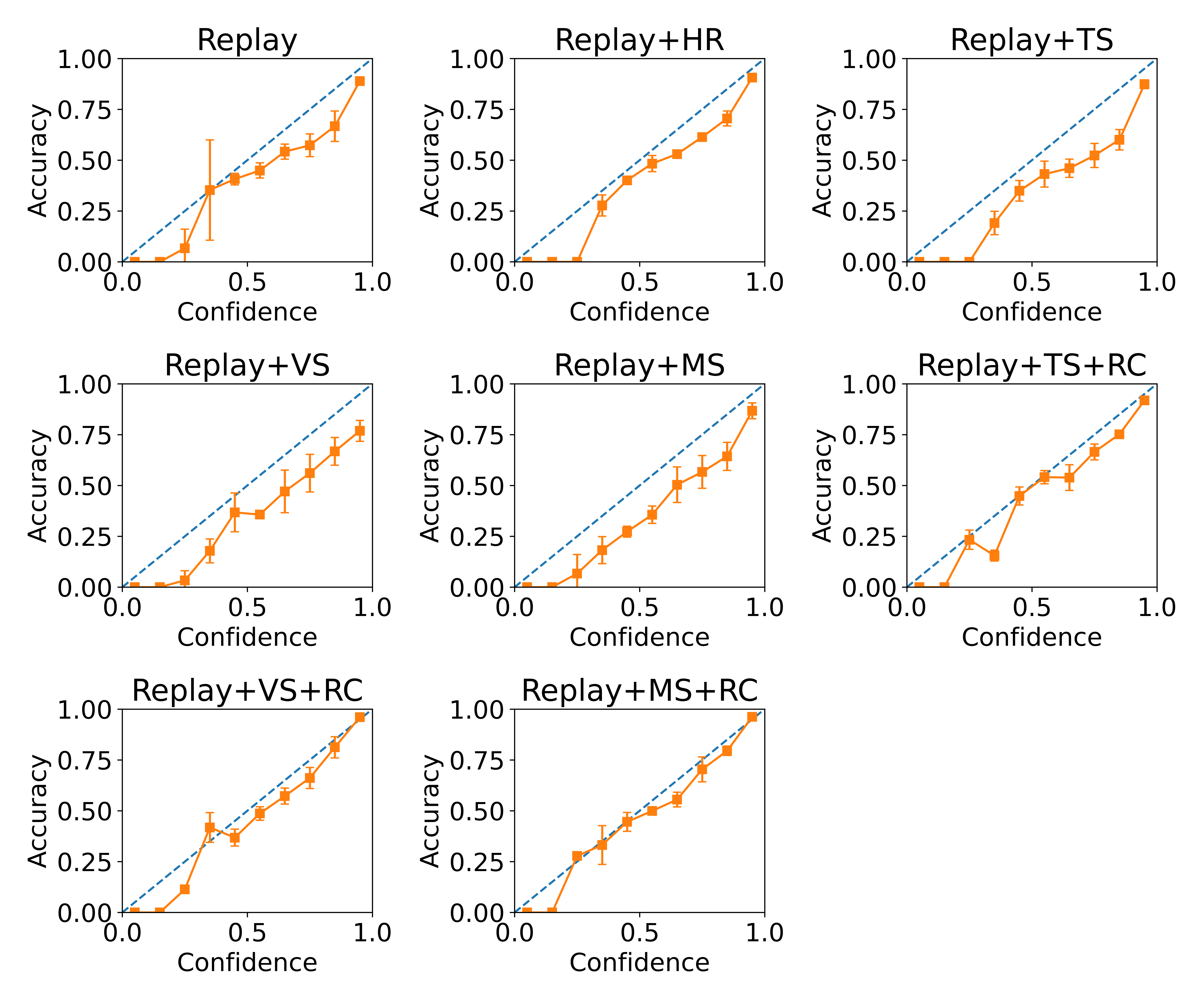}
    \caption{Reliability diagrams for Replay on EuroSAT}
    \label{fig:eurosat-all-replay}
\end{figure}

\begin{figure}
\centering
\includegraphics[width=\linewidth]{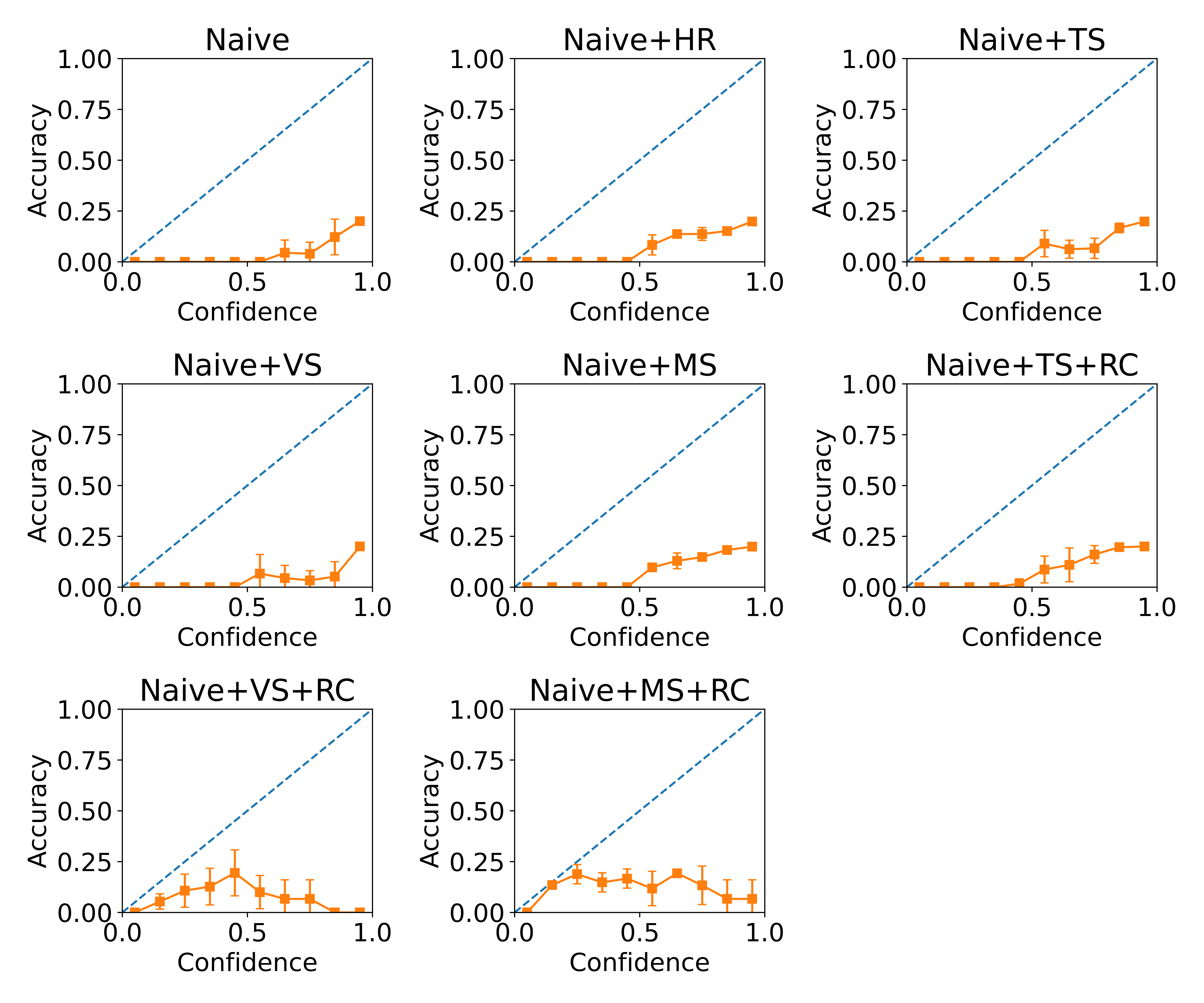}
    \caption{Reliability diagrams for Naive on EuroSAT}
    \label{fig:atari-eurosat-naive}
\end{figure}

\begin{figure}
\centering
\includegraphics[width=\linewidth]{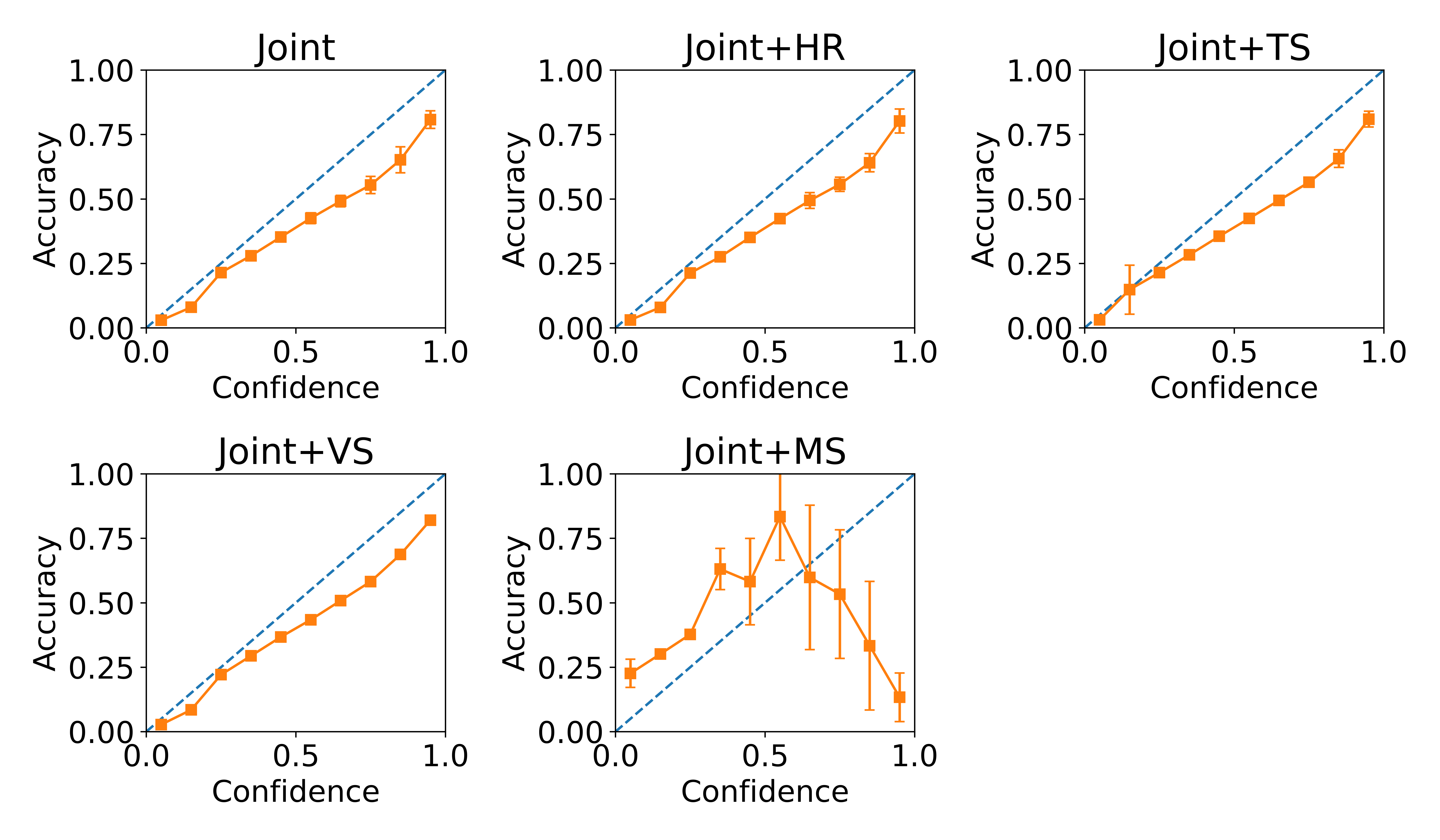}
    \caption{Reliability diagrams for Joint on Atari}
    \label{fig:atari-all-joint}
\end{figure}

\begin{figure}
\centering
\includegraphics[width=\linewidth]{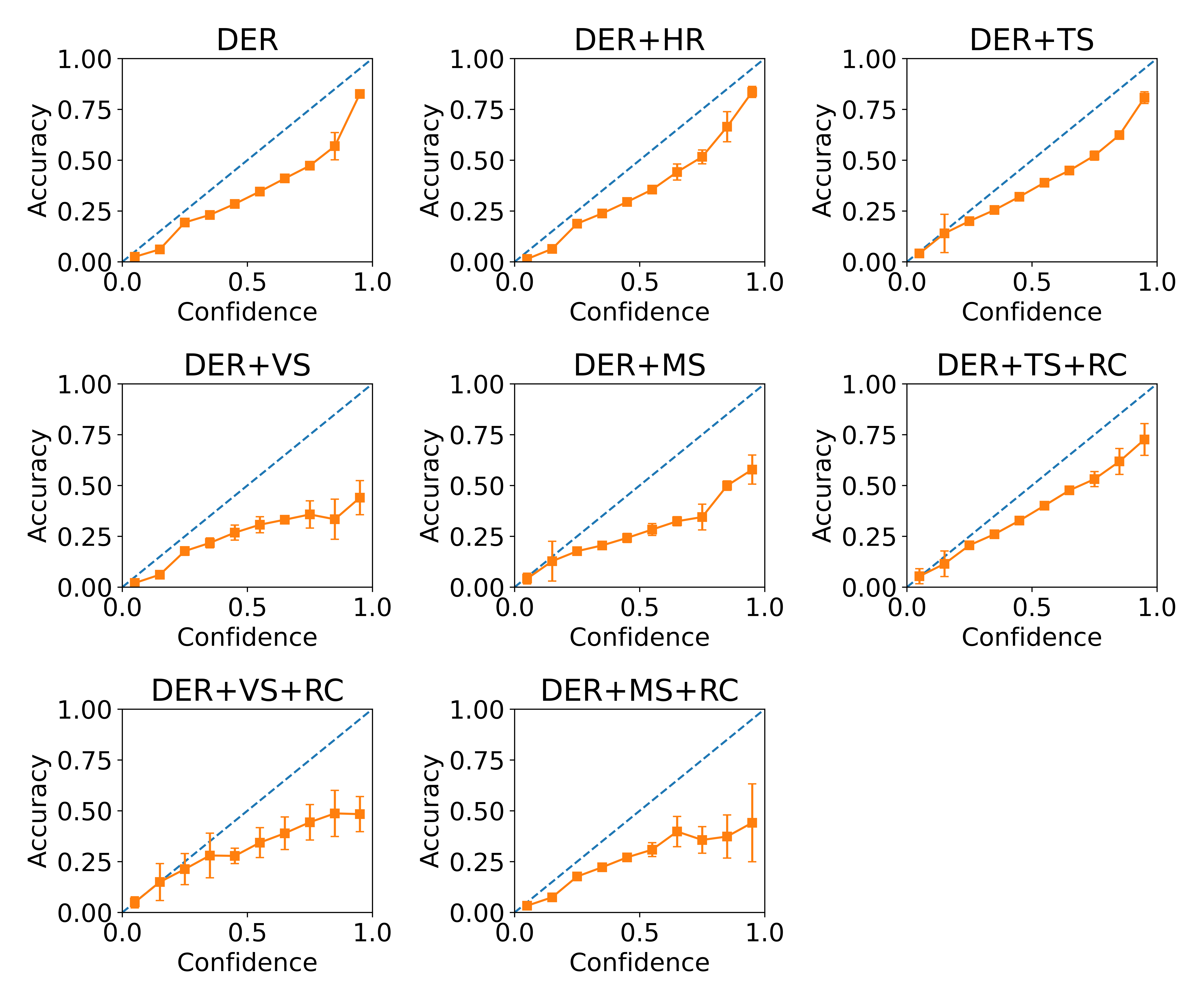}
    \caption{Reliability diagrams for DER++ on Atari}
    \label{fig:atari-all-der}
\end{figure}

\begin{figure}
\centering
\includegraphics[width=\linewidth]{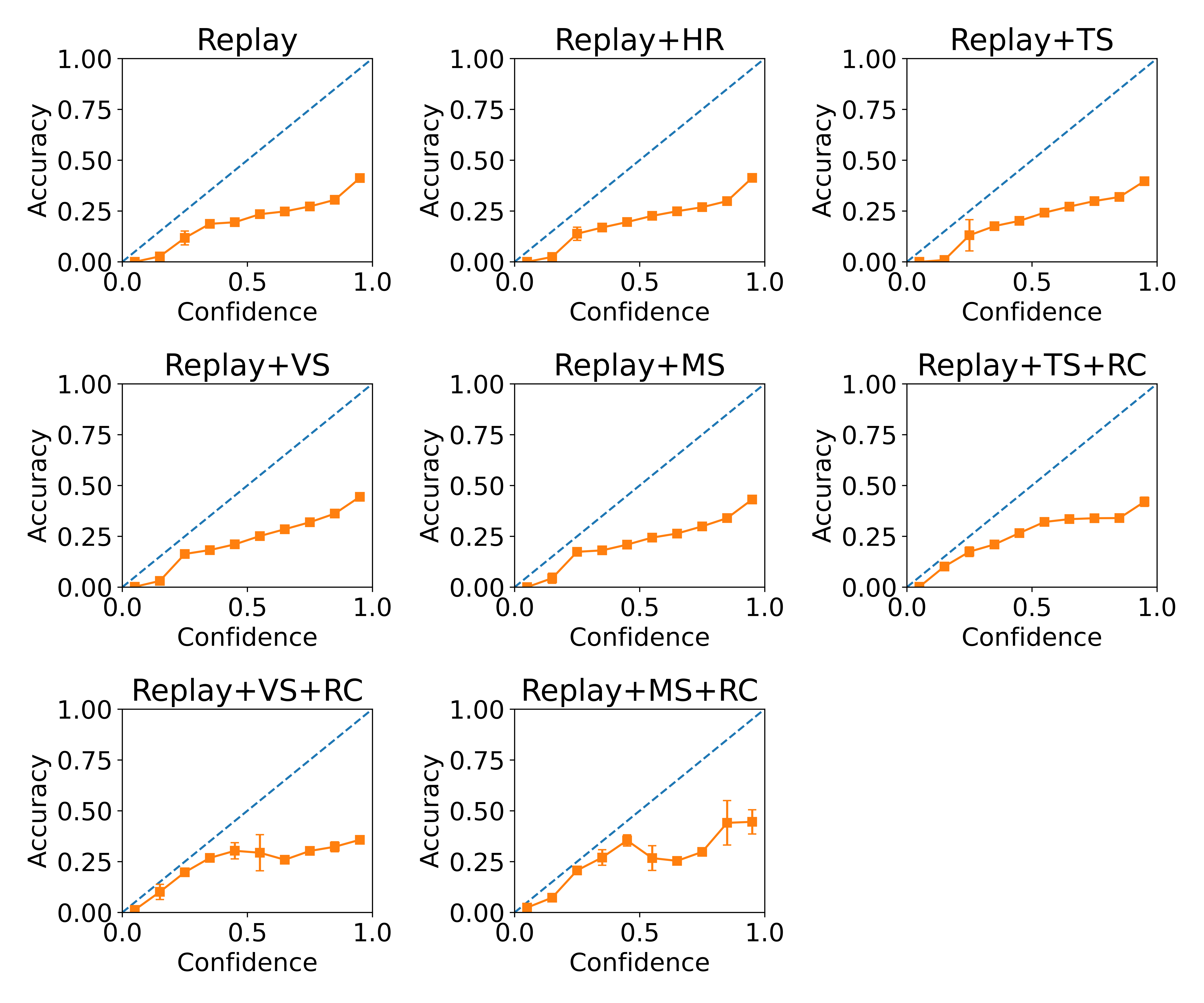}
    \caption{Reliability diagrams for Replay on Atari}
    \label{fig:atari-all-replay}
\end{figure}

\begin{figure}
\centering
\includegraphics[width=\linewidth]{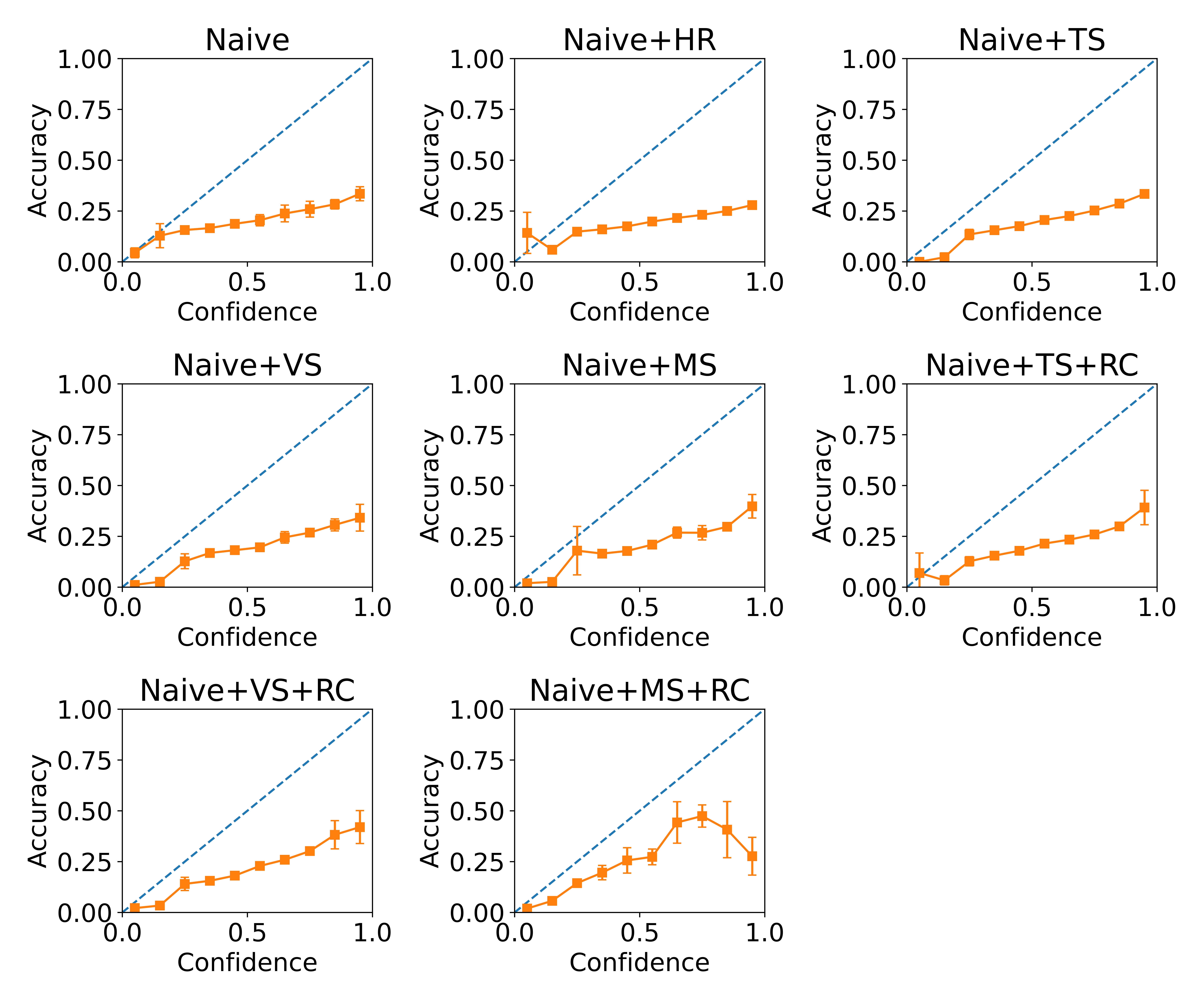}
    \caption{Reliability diagrams for Naive on Atari}
    \label{fig:atari-naive}
\end{figure}

% To split the supplementary pages from the main paper, you can use \href{https://support.apple.com/en-ca/guide/preview/prvw11793/mac#:~:text=Delete%20a%20page%20from%20a,or%20choose%20Edit%20%3E%20Delete).}{Preview (on macOS)}, \href{https://www.adobe.com/acrobat/how-to/delete-pages-from-pdf.html#:~:text=Choose%20%E2%80%9CTools%E2%80%9D%20%3E%20%E2%80%9COrganize,or%20pages%20from%20the%20file.}{Adobe Acrobat} (on all OSs), as well as \href{https://superuser.com/questions/517986/is-it-possible-to-delete-some-pages-of-a-pdf-document}{command line tools}.

\end{document}